\documentclass[journal]{IEEEtran}

\usepackage{times}
\usepackage{epsfig}
\usepackage{graphicx}
\usepackage{amsmath}
\usepackage{amssymb}

\usepackage{multirow}
\usepackage{array}
\usepackage{enumerate}

\usepackage{algorithm}
\usepackage{algorithmic}

\usepackage[pagebackref=true,breaklinks=true,letterpaper=true,colorlinks,bookmarks=false]{hyperref}%,draft
\newcolumntype{C}[1]{>{\centering}p{#1}}

\newcommand{\mytabincell}[2]{\begin{tabular}{@{}#1@{}}#2\end{tabular}}

\DeclareMathOperator*{\argmin}{argmin}

\makeatletter

\newcommand{\Rmnum}[1]{\expandafter\@slowromancap\romannumeral #1@}
\makeatother

\ifCLASSINFOpdf
\else
\fi
\begin{document}
\title{Pyramidal Gradient Matching for Optical Flow Estimation
}

\author{Yuanwei Li% <-this % stops a space
\IEEEcompsocitemizethanks{\IEEEcompsocthanksitem Yuanwei Li is with Chongqing Key Laboratory of Emergency Communication, Chongqing Communication Institute, Chongqing 400035, China (e-mail:lywhbbj@126.com).
}% <-this % stops a space
%Midea Group Co., Ltd., Shunde 410073, China
\thanks{\emph{Corresponding author}: Yuanwei Li.}
}

\maketitle

%\IEEEtitleabstractindextext{
\begin{abstract}
Initializing optical flow field by either sparse descriptor matching or dense patch matches has been proved to be particularly useful for capturing large displacements. In this paper, we present a pyramidal gradient matching approach that can provide dense matches for highly accurate and efficient optical flow estimation. A novel contribution of our method is that image gradient is used to describe image patches and proved to be able to produce robust matching. Therefore, our method is more efficient than methods that adopt special features (like SIFT) or patch distance metric. Moreover, we find that image gradient is scalable for optical flow estimation, which means we can use different levels of gradient feature (for example, full gradients or only direction information of gradients) to obtain different complexity without dramatic changes in accuracy. Another contribution is that we uncover the secrets of limited PatchMatch through a thorough analysis and design a pyramidal matching framework based these secrets. Our pyramidal matching framework is aimed at robust gradient matching and effective to grow inliers and reject outliers. In this framework, we present some special enhancements for outlier filtering in gradient matching. By initializing EpicFlow with our matches, experimental results show that our method is efficient and robust (ranking 1st on both clean pass and final pass of MPI Sintel dataset among published methods).
\end{abstract}

\begin{IEEEkeywords}
Optical flow, gradient image, PatchMatch.
\end{IEEEkeywords}%}

%\maketitle

%\IEEEdisplaynontitleabstractindextext

% \fi
\IEEEpeerreviewmaketitle

\section{Introduction}
\label{Introduction Section}
\IEEEPARstart{A}{ccurate} estimation of optical flow is a useful but challenging topic. The main challenges are large displacements, occlusions, illumination changes, deformations and so on, which are obvious in MPI Sintel and KITTI benchmarks. The traditional approaches, which are based on global energy optimization framework, suffer from these problems more seriously. Recently, many attempts have been done to solve these problems, and an significant way is to initialize optical flow field by either sparse descriptor matching \cite{deepflow} or dense PatchMatch \cite{bao2014cvpreppm, FlowFields,CPMFlow}. Then, sparse-to-dense interpolation and global energy optimization are performed to produce the final optical flow field. Most of the state of the art approaches follow this idea. EpicFlow \cite{EpicFlow} presents a novel method for edge-preserving interpolation from matches. \cite{FlowFields} presents a novel hierarchical correspondence field search strategy which effectively avoids finding outliers.

Under this new framework, good matches provide important guarantee to accurate optical flow estimation with large displacements, deformations and so on. For example, FlowFields \cite{FlowFields} significantly outperforms the original EpicFlow by using its new dense matches instead. In general, sparse-to-dense interpolation requires matches to be as accurate as possible (containing few outliers) and dense enough. Sparse descriptor matches \cite{deepflow} and approximate nearest neighbor fields \cite{random_ann,NNF_kdtrees} (ANNF) are two popular sources used by many optical flow methods. Through selecting points which are confident enough for matching firstly, sparse descriptor matches contain few outliers. However, they are often not dense enough for the following sparse-to-dense interpolation. Compared to sparse descriptor matches, ANNF has the advantage of providing dense matches. As there is no spatial smoothness, ANNF usually contains many outliers. This problem is more pronounced when 1) image patches belong to homogeneous regions or 2) the patch distance between patch pairs of two images is seriously disturbed by illumination changes, deformations and so on. Some techniques (like forward-backward consistency check or its extension \cite{FlowFields}) can be used to remove outliers, but the ultimate goal should be to find as many inliers as possible. For instance, EpicFlow will fail when matches are missing in region whose motion is quite different from its surrounding regions. To make ANNF more robust for optical flow estimation, researchers try many aspects of improvements, such as: 1) designing special search strategy and outlier filter to reject outliers. For example, in the hierarchical matching scheme of \cite{FlowFields}, the hierarchical levels serve as effective outlier sieves; 2) proposing more robust data terms. Instead of patch Euclidean distance, Bao et al. \cite{bao2014cvpreppm} propose an edge aware bilateral data term. \cite{FlowFields} considers two data terms (census transform and patch-based SIFT flow), which are used to handle different benchmarks.

In this paper, we present a pyramidal gradient matching approach that can provide dense matches for highly accurate and efficient optical flow estimation. A novel contribution of our method is that image gradient is used to describe image patches and proved to be able to produce robust matching. Compared to some special features (like SIFT) and patch distance metric, image gradient is much simpler and has better performance with our pyramidal matching framework. Moreover, our work suggests that image gradient is not only efficient and effective but also scalable for optical flow estimation which means we can use different levels of gradient feature. For example, either 6-channel gradients of color image or 2-channel gradients of gray image are effective (without dramatic changes in accuracy) for flow estimation. %So we can obtain different complexity based on different applications.
Especially, our method is even more accurate on KITTI benchmark with fast speed by using 2-channel gradients than 6-channel gradients. Our extreme version (only using direction information of 2-channel gradients of gray image) also has a distinct speed advantage with competitive accuracy. In order to make gradient matches more accurate for optical flow estimation, another contribution is that we uncover the secrets of limited PatchMatch and design a pyramidal matching framework based these secrets. The pyramidal matching framework is aimed at robust matching for large displacements, deformations and so on. Note that, although our pyramidal framework is similar to classical coarse-to-fine framework, our aim (for robust matching) is quite different from coarse-to-fine framework which is mainly for handling large displacements. In our pyramidal framework, we present some enhancements for outlier filtering.

In summary, we make four main contributions:
\begin{enumerate}[(1)]
\item Gradient image is used to provide robust patch matching, and we prove that gradient image is efficient, effective and scalable for optical flow estimation;
\item We uncover the secrets of limited PatchMatch which is a very useful guidance for designing our optical flow algorithm;
\item Based on the secrets of limited PatchMatch, a pyramidal matching framework which is aimed at robust matching is proposed. It is tailored to our gradient matching, especially some enhancements for outlier filtering;
\item We show the effectiveness of our approach by comprehensive experiments on three modern datasets. Especially, among published methods, we obtain best result on MPI Sintel set %and KITTI2015, and second best on KITTI2012
    with competitive speed.
\end{enumerate}
It is indeed incredible that such simple features --- gradient image --- can achieve high accuracy, and so we analysis impacts of gradient image and pyramidal matching framework in Section~\ref{Sec: Effectiveness and scalability} to clear up doubt. It must be noted that the high accuracy of our method is the result of both gradient image and pyramidal matching framework.

The rest of this paper is organized as follows. In Section \ref{Related Work Section}, we review related work. Section \ref{Sec: Our Approach} describes the proposed method which contains gradient image, basic gradient matching, secrets of limited PatchMatch, pyramidal matching framework, outlier filtering and so on. In Section \ref{Sec: Experiments}, we evaluate our method on three modern datasets. Finally, we summarize our work in Section \ref{Conclusions Section}.

%-------------------------------------------------------------------------
\section{Related Work}
\label{Related Work Section}

The classical framework for optical flow estimation is based on variational formulation and the related global energy optimization \cite{hs,middlebury-benchmark}. In order to handle large displacements, coarse-to-fine scheme is widely used. However, such schemes
have intrinsic limitations. The optimization often drops into local optimum and suffers from error propagation effects in multi-scale. Especially, it is unable to preserve motion of structures whose size are too small relative to their motion.

Faced with these problems, many attempts have been made. A straightforward way is to directly search for pixel correspondence without warping and linearization \cite{steinbrucker2009large}. However, it is potentially slow as exhaustive search. Different with exhaustive search, LDOF \cite{LDOF} uses sparse descriptor matching as constraints in its energy-based formulation. Robust keypoints are matched across entire images so that LDOF is able to capture large displacements. MDP-Flow \cite{MDP-Flow2} fuses the flow propagated from the coarser level and the sparse SIFT matches to improve the initial flow at each level. In \cite{deepflow}, Weinzaepfel et al. propose a descriptor matching algorithm (called DeepMatch), which is tailored to the optical flow estimation and can produce dense correspondence field efficiently.

In addition to sparse descriptor matching, ANNF has also been proven to be an effective information source for large displacements. Especially, a seminal work called PatchMatch \cite{random_ann} brings a lot of interests to ANNF-based optical flow estimation because of its computational efficiency. One of the first work that employs ANNF for optical flow estimation is \cite{cvpr-Chen-Large}. It uses ANNF to generate an initial dense but noisy matching which is refined through motion segmentation. Lu et al. \cite{lu2013patch} propose a variant of PatchMatch which uses superpixels to obtain edge aware field. Bao et al. \cite{bao2014cvpreppm} bring in spatial smoothness by increasing the patch size and preserve motion edges by a special matching cost function. In addition, a self-similarity propagation scheme is used to match big patches fast. In \cite{FlowFields}, Bailer et al. propose a novel purely data based hierarchical search strategy which is able to find most inliers and avoid finding outliers effectively.

Regrading sparse-to-dense interpolation, EpicFlow \cite{EpicFlow} is a seminal work. Based on a novel edgeaware geodesic distance, it provides an effective way to fill the gaps between matches and the final dense flow. For instance, both the original EpicFlow (using matches in \cite{deepflow,KD-Trees} as inputs) and FlowFields+EpicFlow \cite{FlowFields} achieve high accuracy on MPI Sintel benchmark \cite{MPI-benchmark}.

There are also some works that share the idea of utilizing gradient information with our method. Gradient constancy assumption has been used in \cite{Tistarelli1994,Uras1988} to deal with the aperture problem. Brox and Malik \cite{HAOF} extend the brightness constancy assumption by a gradient constancy assumption to enhance the robustness against gray value changes. In order to use the more fitting constraint for different areas, a binary weight map is used to switch between brightness constraint and gradient constraint in \cite{MDP-Flow2}. In LDOF \cite{LDOF}, HOG features (based on gradient information) are used to produce dense matches for the energy-based formulation.

\section{Our Approach}
\label{Sec: Our Approach}
In this section, we give a detailed introduction to our pyramidal gradient matching approach from following aspects: gradient image, basic gradient matching, secrets of limited PatchMatch, pyramidal matching, outlier filtering and the scalability of gradient image. %\textbf{\emph{gradient image} (the form defined in this paper for using image gradient in patch matching)}.
A novel idea of our approach is to use image gradient information in the form of \emph{gradient image} (defined in Section~\ref{Sec: Image Gradients}) to describe image patches. Based on this idea, we first describe a basic gradient matching method in Section~\ref{Sec: gradient matching}. To obtain dense correspondence patches described by gradient image,  basic gradient matching adopts limited PatchMatch as the basic matching method. A direct use of basic gradient matching is far from the requirement of highly accurate optical flow estimation. Based on a thorough analysis of limited PatchMatch (Section~\ref{Sec: Secret of Limited PatchMatch}), we propose a special pyramidal matching framework which is tailored to gradient matching (Section~\ref{Sec: Hierarchical Matching}). Through pyramidal matching, our approach is able to find more inliers and reject outliers effectively. Especially, we introduce some special enhancements for outlier filtering (Section~\ref{Sec: Outlier Filtering}). Furthermore, as another advantage, we discuss the scalability of gradient image in Section~\ref{Sec: Flexibility of Gradients}. Finally, Section~\ref{Sec: Sparsification and Dense} describes the EpicFlow-based sparse-to-dense interpolation. %\textbf{which partly shares the idea with \cite{FlowFields}}

\subsection{Gradient Image}
\label{Sec: Image Gradients}

\begin{figure}[t]
\begin{center}
   \includegraphics[width=0.99\linewidth]{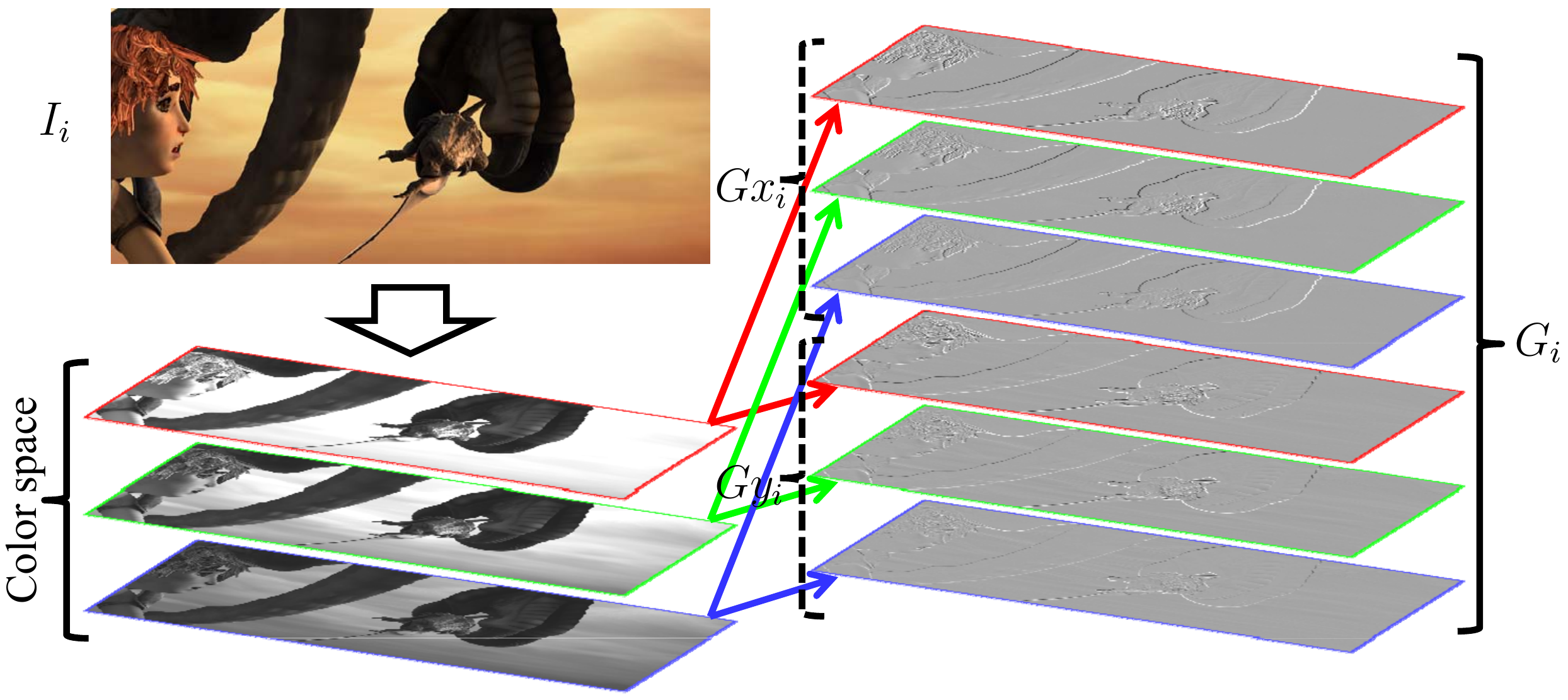}
\end{center}
   \caption{A visual representation of the computation of gradient image. In this figure, the color space is RGB.}
\label{Fig: Gradients computation by Sobel}
\end{figure}

The original PatchMatch \cite{random_ann} defines matching cost as patch Euclidean distance in CIELab color space. In order to make PatchMatch more robust for optical flow estimation, an effective improvement is to modify matching cost. For instance, SIFT flow and census transform are used in \cite{FlowFields}, and Bao et al. \cite{bao2014cvpreppm} propose an edge-preserving matching cost. In this paper, we propose to use image gradient as basic feature for patch matching.

Image gradient, or image derivative, is one of the most common features which is very simple for computing. It has been used to deal with brightness or color constraint disturbance (such as illumination change) \cite{HAOF,MDP-Flow2}. Compared to gray and color information, gradient vector of one pixel is able to hold structure information of a local region. For one pixel, its gray or color vector is 1-dimension or 3-dimension, while, its gradient vector is 2-dimension or 6-dimension which can be decomposed into direction and magnitude. In view of these, we try to handle challenging cases (like large displacements, scale/rotational deformations) by directly using image gradient without other features.

Given a source image $I_i$, its horizontal gradient $Gx_i$ and vertical gradient $Gy_i$
can be computed by using Sobel operator
\begin{equation}
\left\{
\begin{split}
Gx_i= K_{x}(S)*I_i
\\
Gy_i= K_{y}(S)*I_i\end{split}
\right.
,
\label{Eq: Gradients computation by Sobel}
\end{equation}
where $*$ here denotes the 2-dimensional convolution operation. $K_{x}(S)$ and $K_{y}(S)$ are Sobel kernels in two directions, and $S\in \{3,5,7,...\}$ is the kernel size.% For the sake of clarity, an example of $K_{x}(S)$ used in our approach is given in Table~\ref{Tab: an example of Sobel kernels}.

%\begin{table}[t]
%\begin{center}
%\caption{An example of Sobel kernels. This table give two kernels in  $x$ direction whose size are $S=3$ and $S=5$ $K_{x}(S)$ respectively.}
%\label{Tab: an example of Sobel kernels}
%\includegraphics[width=0.7\linewidth]{sobel_kernels.pdf}
%\end{center}
%\end{table}

Now, we define $G_i$ as \emph{gradient image} of $I_i$. The corresponding pixel $G_i(p)$ of $G_i$ at position $p=(x,y)$  is computed as follows
\begin{equation}
G_i(p)=[Gx_i(p),~Gy_i(p)]^T.
\label{Eq: Define of gradients image}
\end{equation}

Eq.~(\ref{Eq: Define of gradients image}) shows that each pixel of $G_i$ is described by the gradient in $x$ and $y$ direction of its corresponding pixel in $I_i$. A visual representation of the computation of gradient image $G_i$ is given in Figure~\ref{Fig: Gradients computation by Sobel}. In this figure, $I_i$ is supposed to be color image in RGB color space (3-channel). Therefore, $Gx_i$ and $Gy_i$ are 3-channel respectively, and $G_i$ is 6-channel.

\subsection{Basic Gradient Matching}
\label{Sec: gradient matching}

Gradient matching is to find a correspondence field between two gradient images. Our basic gradient matching follows PatchMatch \cite{random_ann}. The basic idea of PatchMatch is to generate a random correspondence field firstly and perform an iterative process of improving it. In each iteration, it contains two types of operations: propagation and random search. Propagation is to propagate good guesses to neighboring pixels, and random search is to find better guesses by trying several random guesses near current guesses.

A simple way to modify PatchMatch is to replace CIELab color image with our gradient image, however, we find it is too rough to meet the requirement of highly accurate optical flow estimation. When facing challenging cases, it often leads to too sparse matches and a lot of outliers. In this paper, following the definition in \cite{FlowFields}, we regard that outliers can be divided into two categories : \emph{mild outliers} and \emph{resistant outliers}. Here, mild outliers are these that may be found or pass outlier filter accidentally so that we can reject them by performing special matching scheme (like the hierarchical matching in \cite{FlowFields}% and our pyramidal matching
), multiple filtering and so on. While, resistant outliers are these that can not be removed as their matching cost are low enough so that we even believe they are truth matches if no more information (like structure information in a larger region) is introduced. As said in \cite{FlowFields}, the secret to reject resistant outliers is to avoid finding them. %, so we only consider outlier filtering of mild outliers here.
For mild outliers, although some techniques %such as forward-backward consistency check
can be used to remove them, they also leave gaps in the field. More seriously, too many gaps over a region may lead to motion loss in the whole region (such as thin structures). Therefore, not only removing outliers by outlier filtering, we need to find most inliers instead of outliers to the maximum possible extent.

For this purpose, in addition to gradient image, our basic gradient matching also adopts the idea of limiting random search distance in PatchMatch (we call it \emph{limited PatchMatch}), which has been proved to be powerful in optical flow estimation \cite{FlowFields, CPMFlow}. The reason why outliers are found by PatchMatch is that these outliers are better than ground truth in the search region, and so limited PatchMatch is an effective way to reject outliers as disturbance terms are nonexistent in a small limited search region. A precondition for finding inliers and rejecting outliers by limiting search distance is that ground truth is around the search center which is determined by the initial field. Therefore, we need to provide good initial field for limited PatchMatch and enhance outlier filtering. In the rest of this section, we will detail our basic gradient matching (combine gradient image and limited PatchMatch) without consideration of initial field and outlier filtering. In next Section, we attempt to uncover what has made limited PatchMatch powerful through a thorough analysis, and explain designing ideas of our pyramidal matching framework based on these analysis. Section~\ref{Sec: Hierarchical Matching} describes how to grow inliers and reject outliers with a pyramidal matching framework in detail.

Suppose $I_1$ and $I_2$ are two input images that we need to estimate the optical flow between them, the gradient images $G_1, G_2$ of two input images can be computed according to Section~\ref{Sec: Image Gradients}. Then, we directly compute correspondence field $F$ based on $G_1, G_2$ instead of $I_1, I_2$. If we denote a square patch of $G_i$ with patch size $(2r+1)\times (2r+1)$ centered at position $p=(x,y)$ by $P_i(p)$, $F$ gives the corresponding patch $P_2(F(p))$ in $G_2$ for each patch $P_1(p)$ in $G_1$. %Therefore, the goal of our gradient matching is to find the corresponding field $F$.Here, $P_r(x,y,G)$ is used to denote a square patch of \emph{gradients images} $G$ with patch size $(2r+1)\times (2r+1)$ centered at position $(x,y)$. The goal of gradient matching is to find the corresponding gradient patch $P_r(x_b,y_b,G_2)$ in the second image for a patch $P_r(x_a,y_a,G_1)$ in the first image.
As defined in Section~\ref{Sec: Hierarchical Matching}, some pixels of $F$ are initialized with good guesses around ground truth, and the remaining pixels are marked as $uninitialized$. %In order to solve $F$, we firstly randomly initialize $F$. In addition, some special techniques (such as the kdtree based initialization step of \cite{KD-Trees}) are also available.
In order to improve $F$, we iteratively perform propagation and limited random search operations. Given two positions: $p_a=(x_a,y_a)$ in $G_1$ and $p_b=(x_b,y_b)$ in $G_2$, the matching cost between $P_1(p_a)$ and $P_2(p_b)$ is defined as
\begin{equation}
D_r(p_a,p_b)=\sum\limits_{
^{~p_o=(x_o,y_o)}
_{|x_o|\leqslant r, |y_o|\leqslant r}}
{||G_1(p_a+p_{o})-G_2(p_b+p_{o})||^2}.
\label{Eq: Define of patch distance}
\end{equation}
Eq.~\ref{Eq: Define of patch distance} shows that we compute $L_2$ distance to measure the difference between two gradient patches.% It is computational efficient, especially for the propagation operation.

In propagation, the new corresponding position $F(p_a)$ of $P_1(p_a)$ is improved as
\begin{equation}
F(p_a)= \argmin \limits_{p_p \in \Omega} (D_r(p_a,p_p)),
\end{equation}
$$\Omega=\{F(p_a), F(p_a-\Delta_x)+\Delta_x, F(p_a-\Delta_y)+\Delta_y \}.$$
$\Delta_x$ and $\Delta_y$ determine the propagation direction. For instance, the changing order of $\Delta_x$ and $\Delta_y$ is
\begin{equation*}
\small{
\left\{
\begin{split}
\Delta_x:(1,0)
\\
\Delta_y:(0,1)
\end{split}
\rightarrow
\begin{split}
(-1,0)
\\
(0,-1)
\end{split}
\right.
}
\end{equation*}
in the original PatchMatch, while FlowFields adopts
\begin{equation*}
\small{
\left\{
\begin{split}
\Delta_x:(1,0)
\\
\Delta_y:(0,1)
\end{split}
\rightarrow
\begin{split}
(-1,0)
\\
(0,-1)
\end{split}
\rightarrow
\begin{split}
(-1,0)
\\
(0,1)
\end{split}
\rightarrow
\begin{split}
(1,0)
\\
(0,-1)
\end{split}
\right.
.}
\end{equation*}
We change $\Delta_x$ and $\Delta_y$ in the same order as FlowFields, however, we find that the changing order of $\Delta_x$ and $\Delta_y$ has very little impacts on the accuracy of our approach. Note that, as some pixels of $F$ are $uninitialized$, each element of $\Omega$ must be checked before propagating. For example, element $F(p_a-\Delta_x)+\Delta_x$ is ignored if $F(p_a-\Delta_x)$ is $uninitialized$.

The propagation of patch $P_1(p_a)$ is followed by a limited random search to improve $F(p_a)$ further. As PatchMatch \cite{random_ann}, we try to find better corresponding patch from a sequence of candidates at an exponentially decreasing distance from $F(p_a)$ as
%$$p_i=F(p_a)+ \frac{R_iW}{2^i},$$
\begin{equation}
F(p_a)= \argmin \limits_{p_s\in \Psi} (D_r(p_a,p_s)),
\label{Eq: Limiting search}
\end{equation}
$$\Psi=\{F(p_a)\}\cup\{p|p=F(p_a)+ \lfloor\frac{R_iW}{2^i}\rfloor\},$$
where $\lfloor z\rfloor$ is to round to the largest integer that does not exceed $z$, $R_i$ is a random uniformly distributed offset in $[-1,1]\times[-1,1]$, and $i=0,1,...,\lfloor \log_2{W} \rfloor$. Base on Eq.~(\ref{Eq: Limiting search}), the idea of limited random search is implemented by assigning a small value to $W$. %($W=2$ in our experiments).
As the definition of $\Psi$, we do not consider subpixel accuracy in our gradient matching.
\subsection{Secrets of Limited PatchMatch}
\label{Sec: Secret of Limited PatchMatch}

\begin{figure}[t]
\begin{center}
   %\includegraphics[width=0.4\linewidth]{grts_channel1.pdf}
   \includegraphics[width=0.99\linewidth]{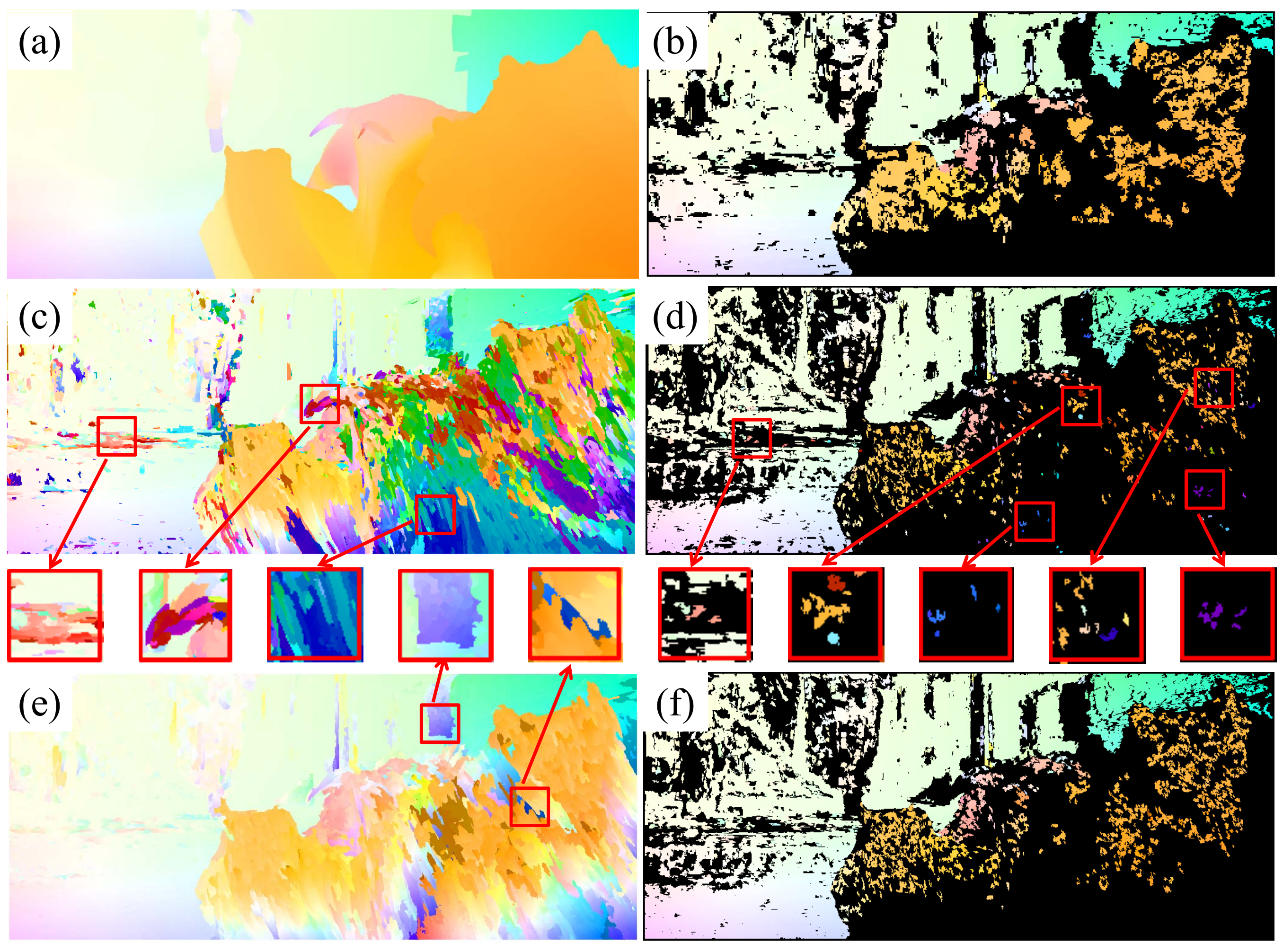}
\end{center}
   \caption{An example of the power of our basic gradient matching. (a) Ground
truth. (b) is the initial correspondence field for (c) and (e), which is obtained by pyramidal gradient matching as described in Section~\ref{Sec: Hierarchical Matching}. (c) is the correspondence field without limited random search, and (d) is its filtered field. (e) is the correspondence field with limited random search (our basic gradient matching), and (f) is its filtered field. Note that, instead of CIELab color image, all results are computed with gradient image.}
\label{Fig: markert5-frame15}
\end{figure}

\begin{figure}[t]
\begin{center}
   %\includegraphics[width=0.4\linewidth]{grts_channel1.pdf}
   \includegraphics[width=0.99\linewidth]{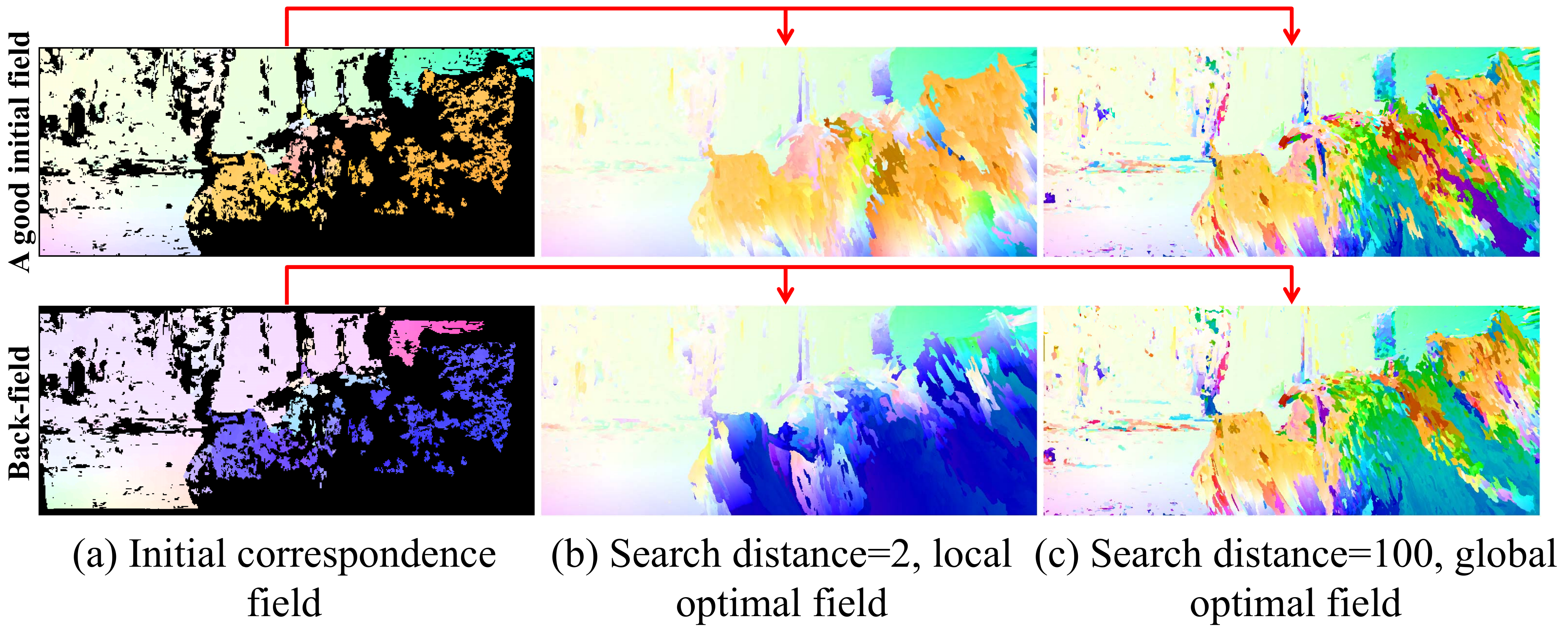}
\end{center}
   \caption{An example of local/global optimal field. (a) is two initial correspondence fields: the above field is a good initial field which is close to the ground truth field (see Figure~\ref{Fig: markert5-frame15}(a)), and the under field is the back-field which is obtained by reversing the above field. Based on the initial field in respective rows, (b) is two local optimal fields obtained by limiting random search distance and (c) is two global optimal fields with a large enough random search distance.}
\label{Fig: local_global_optimal}
\end{figure}

\begin{figure}[t]
\begin{center}
   %\includegraphics[width=0.4\linewidth]{grts_channel1.pdf}
   \includegraphics[width=0.99\linewidth]{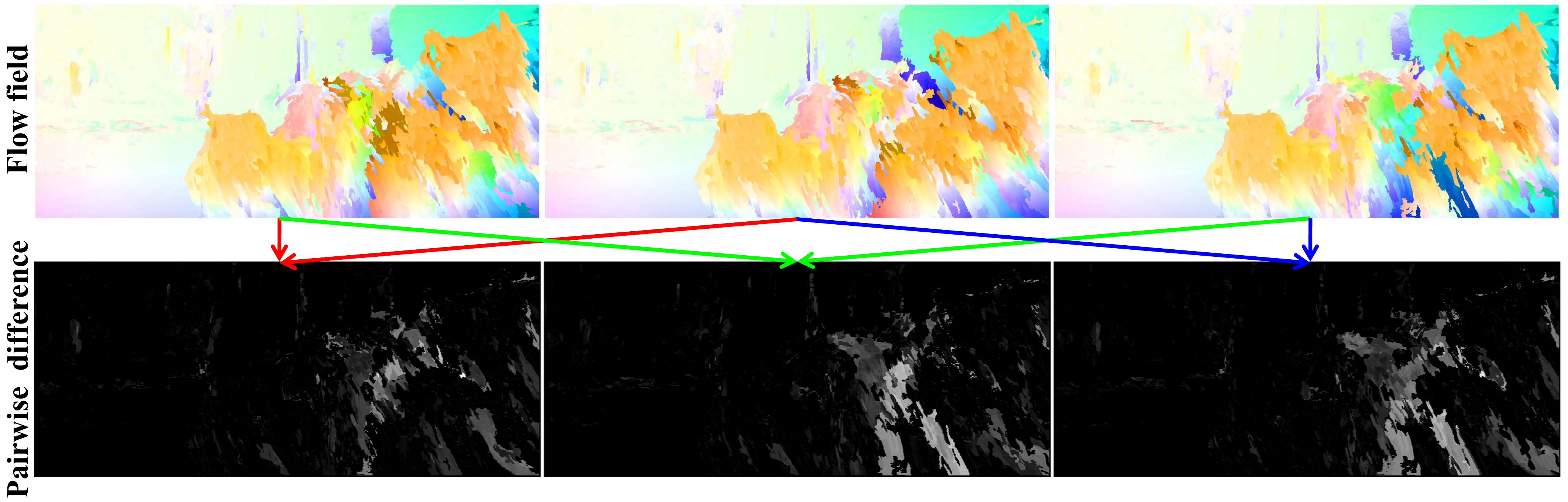}
\end{center}
   \caption{An example of stable/unstable local optimal flow. The first row presents three local optimal fields based on the same initial flow field. Note that, the only difference among three fields is their random seeds for PatchMatch's random search step. The second row presents the difference (endpoint error) between every two fields.}
\label{Fig: stable_unstable}
\end{figure}

In this section, through a thorough analysis, we try to uncover the secrets of limited PatchMatch which is a very useful guidance for designing our optical flow algorithm.

The basic idea of limited PatchMatch is to limit the search distance in the random search step of PatchMatch. Given a search distance, we believe that the original PatchMatch is effective to find correspondence patches with low enough match cost in the search space determined by this distance. %Or in other words, PatchMatch is a effective optimization method .
Therefore, as shown in Figure~\ref{Fig: markert5-frame15}(c,d), the original PatchMatch (but using gradient image instead of CIELab color image) may generate a lot of outliers (especially the five marked regions in Figure~\ref{Fig: markert5-frame15}(d)) even if the initial field is good (see Figure~\ref{Fig: markert5-frame15}(b)).
The reason is that the matching cost of ground truth is not small enough compared to these outliers introduced by searching in a large region. In limited PatchMatch, we try to find matches with smallest matching cost in a small region guaranteed by a small random search distance. In other words, limited PatchMatch is to find \textit{local optimal field} around initial field, while the original PatchMatch is to find \textit{global optimal field}. To further explain the meaning of ``local/global optimal field'', we respectively run limited PatchMatch (the generated field Figure~\ref{Fig: local_global_optimal}(b) is called local optimal field) and original PatchMatch (the generated field Figure~\ref{Fig: local_global_optimal}(c) is called global optimal field) based on two reciprocal initial fields. As shown in Figure~\ref{Fig: local_global_optimal}(a), the first row is based on a good initial field, while the second row is based on the back-field. Under the condition that two initial fields are quite different, the results show that two local optimal fields are quite different (or similar to their corresponding initial field), while two global optimal fields are quite similar.% So limited PatchMatch is effective to generate a continuous and local optimal field around the initial field., and the local optimal field around good initial field is probably

Further, we find that the optimization ability of limited PatchMatch is still powerful even searching in limited region. As shown in Figure~\ref{Fig: markert5-frame15}(e,f), limited PatchMatch is very powerful to find local optimal motion which may be quite different from the initial flow even if the search distance is very small ($=2$ in this figure). For example, it is able to generate a continuous field with sparse initial field, moreover, some outliers (the two marked regions in Figure~\ref{Fig: markert5-frame15}(e)) are also found even if no similar initial seeds (see Figure~\ref{Fig: markert5-frame15}(b)) are given as its powerful optimization ability. %The comparison between Figure~\ref{Fig: markert5-frame15}(c,d) and Figure~\ref{Fig: markert5-frame15}(e,f) shows that
So limited PatchMatch is effective to enlarge inlier regions around a good initial field, and meanwhile avoid finding outliers which are hard to remove.

Thirdly, for each flow vector in local optimal field, we divide it into two kinds: \textit{stable local optimal flow} and \textit{unstable local optimal flow}. As shown in Figure~\ref{Fig: stable_unstable}, with same initial field but different random seeds for PatchMatch's random search step, three resulting local optimal fields are not exactly the same --- quite different in the right-bottom challenging region but nearly same in remaining region. Following our definition, the right-bottom region is unstable local optimal flow and the remaining region is stable local optimal flow. A positive effect of this definition is that inliers usually belong to stable local optimal flow and outliers usually belong to unstable local optimal flow, so we can remove outliers with the help of this instability. Especially, this instability is more meaningful when other differences among multiple PatchMatch iterations --- unlike the only random seeds difference in Figure~\ref{Fig: stable_unstable} --- are introduced.

Based on the above analysis, we sum up following four secrets or key points of limited PatchMatch:
\begin{enumerate}[(1)]
\item Limited PatchMatch is an effective way to find local optimal field around initial field, which grow inliers effectively.
\item Limited PatchMatch has a powerful local optimization ability.
\item Local optimal field can be divided into stable local optimal flow and unstable local optimal flow, and outliers usually belong to unstable local optimal flow.
\item Local optimal field obtained by limited PatchMatch based on a good initial field usually is a well estimation of ground truth.
\end{enumerate}

\begin{table}[t]
\begin{center}
\caption{Impacts of different aspects of our method on estimation accuracy. AEE of our complete PGM-C (first row) serves as the baseline for performance comparison. For other rows, we modify PGM-C in one respect as described in the first column and report AEE and AEE increment with respect to complete PGM-C. Note that, AEE is reported on $\frac{1}{3}$ final pass of MPI training set.}
\label{Tab: Secrets of limited PatchMatch}
\small{
\begin{tabular}{c|c|c}
\hline
Method description & AEE & Increment\\
\hline
\hline
Our complete PGM-C (Section~\ref{Sec: Flexibility of Gradients})
     & 3.478	& ---\\
\hline
Without refinement of initial field  & 3.629	& 4.34\%\\
Propagate both inliers and outliers   & 3.636	& 4.54\%\\
Without outlier record propagation   & 3.571	& 2.67\%\\
%Without color space difference (Table~\ref{Tab: settings of gradient pyramid})  & 3.490	& 0.35\%\\
%Add $[-1,1]$ random noise in correspondence field propagation  & 3.484	& 0.17\%\\
\hline
\end{tabular}
}
\end{center}
\end{table}

Now, we try to construct an effective matching framework for optical flow estimation based on these secrets. The key point is to provide a good initial field for limited PatchMatch at the full resolution level, so the resulting local optimal field usually is a well estimation of ground truth. To obtain good initial field, we must reject outliers as completely as possible. In \cite{FlowFields,CPMFlow}, hierarchical architectures with one-way propagation from top to bottom are used for outlier filtering. Their core idea, as said in \cite{FlowFields}, is that ``\emph{Propagation and random search (with small enough $R$) are usually too local in flow space to introduce new outliers, while propagations of lower hierarchy levels are likely to remove most outliers persisting in higher levels, since resistant outliers are often not resistant on all levels.}'' Compared to \cite{FlowFields,CPMFlow}, our method mainly differ in following aspects:
%\begin{enumerate}[1)]
\\$\bullet$ As shown in Figure~\ref{Fig: outline of hierarchical matching}, we design two kinds of hierarchy levels for ``refinement of initial field'' (iteratively propagating between two specific levels) and ``propagation of the refined field'' respectively, while there is no clear distinction in \cite{FlowFields,CPMFlow} where refinement and propagation are mixed together. Experimental verification is given in Table~\ref{Tab: Secrets of limited PatchMatch}: row ``Our complete PGM-C'' vs. row ``Without refinement of initial field''. Our viewpoint that good initial field is so important for limited PatchMatch motivates us to introduce special ``refinement of initial field''. Through refinement of initial field, we can remove outliers more completely than the one-way propagation from top to bottom in \cite{FlowFields,CPMFlow}. One worry is that the refined field may be too sparse, however, as said in Secret~(2), limited PatchMatch --- boosted by special levels for ``propagation of the refined field'' --- is powerful enough to generate local optimal field around the good initial field.
\\$\bullet$ Our ``Correspondence field propagation'' operation (blue arrow in Figure~\ref{Fig: outline of hierarchical matching}) is different. We perform forward-backward consistency check at all levels and only propagate flow inliers to following level, while \cite{FlowFields,CPMFlow} only perform check on the bottom level and propagate all flow vectors (both inliers and outliers) to following level. Experimental verification is given in Table~\ref{Tab: Secrets of limited PatchMatch}: row ``Our complete PGM-C'' vs. row ``Propagate both inliers and outliers''. The ability of our different ``Correspondence field propagation'' that can remove outliers more completely is also closely related to Secret~(2). If propagating outliers to following level (that is, these outliers are used as initial flow for limited PatchMatch at following level), the powerful local optimization ability of limited PatchMatch is probably to generate continuous motion regions around these outliers. Therefore, new outliers are introduced if some pixels in these motion regions pass consistency check even if they can not pass consistency check at previous level. In our method, by only propagating inliers, we can remove more outliers. Similarly, the powerful local optimization ability ensures that we can preserve inlier motions with a sparse initial field.
\\$\bullet$ Except propagating correspondence field between levels, we introduce ``Outlier record propagation'' operation (red arrow in Figure~\ref{Fig: outline of hierarchical matching}). That is, whether a pixel is an outlier depends on not only forward-backward consistency check result ($yes$/$no$) at current level, but also its outlier record ($yes$/$no$) from previous levels. This pixel is regarded as an outlier when either is $yes$, and then its new outlier record will be propagated to following level. Experimental verification is given in Table~\ref{Tab: Secrets of limited PatchMatch}: row ``Our complete PGM-C'' vs. row ``Without outlier record propagation''. Our motivation of introducing ``Outlier record propagation'' is Secret~(3). As said in Secret~(3), outliers usually belong to unstable local optimal flow, so we can identify some hard outlier pixels which may pass consistency check at some levels by considering its outlier record at multi-levels. Once a pixel is judged as outlier at one level, we regard it as an outlier at following levels. Note that, as shown in Figure~\ref{Fig: outline of hierarchical matching}, except one blue arrow, other blue/red arrows come in pairs. This single blue arrow means that we do not propagate outlier record between two connection levels between ``refinement of initial field'' and ``propagation of the refined field'', and the reason is that propagating outlier record further may make correspondence field too sparse. Therefore, ``propagation of the refined field'' only receives the filtered initial field from ``refinement of initial field'' and ignores outlier record.
%\end{enumerate}

As show in Table~\ref{Tab: Secrets of limited PatchMatch}, our matching framework, especially the above three points, can reduce estimation error significantly. The details of our matching framework (called \textit{pyramidal matching}) will be given in the next section.

\subsection{Pyramidal Matching}
\label{Sec: Hierarchical Matching}

\begin{figure}[t]
\begin{center}
   \includegraphics[width=0.99\linewidth]{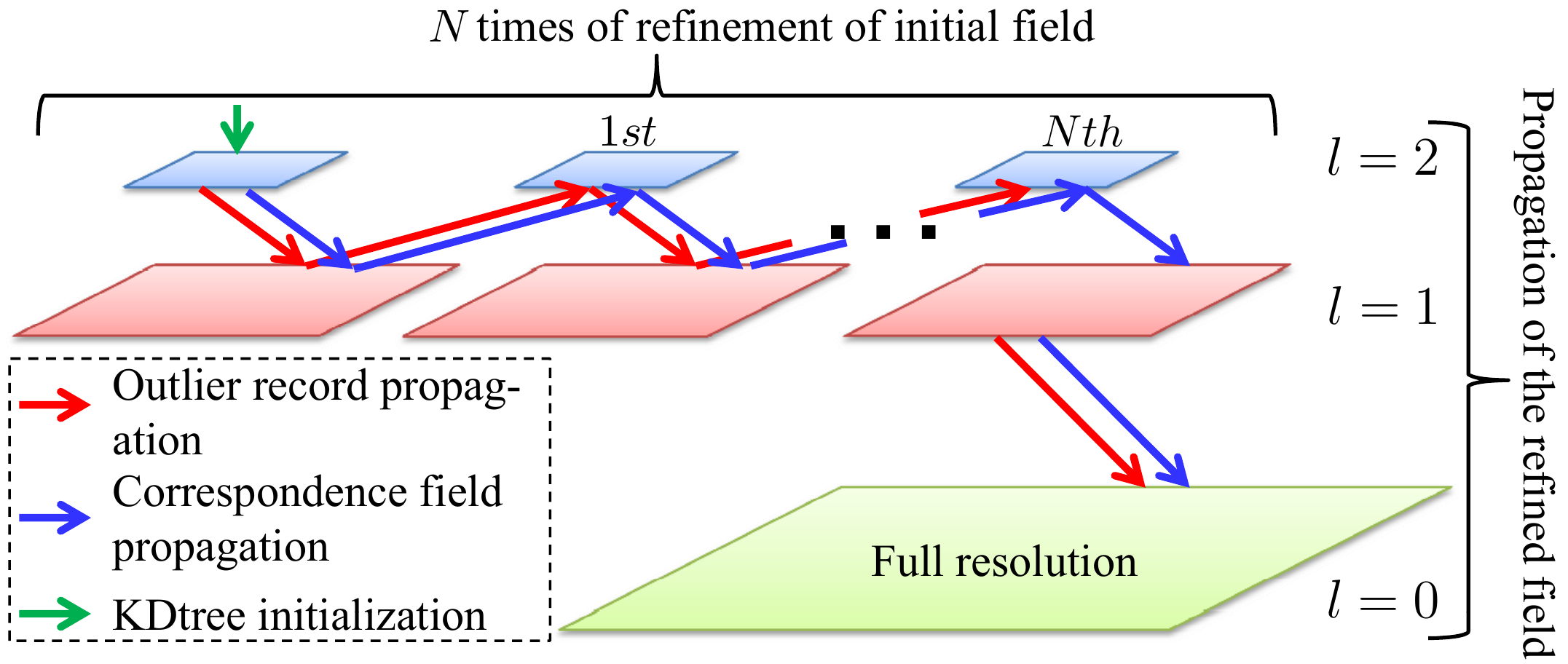}
\end{center}
   \caption{The outline of our pyramidal matching. The pyramidal matching consists of two steps: refinement of initial field and propagation of the refined field, which involve three different operations (represented by arrows in different colors) when propagating field between neighboring levels.}
\label{Fig: outline of hierarchical matching}
\end{figure}

Our pyramidal matching framework is designed to filter more outliers and provide good initial field for limited PatchMatch at the full resolution level. The core idea of our pyramidal matching framework has been discussed in Section~\ref{Sec: Secret of Limited PatchMatch}, and this section will give a detailed description. Figure~\ref{Fig: outline of hierarchical matching} is the outline of our framework, which consists of two steps: 1) We refine the initial field by propagating correspondence field and outlier record between two specific levels iteratively. Through the refinement of initial field, most of outliers are removed; 2) The refined initial field (without previous outlier record) are propagated to the full resolution level level-by-level.

We define that $G^l_i$ ($l=0,1,...,L-1$) is the $l$th level of \emph{gradient pyramid} $\{G^0_i, G^1_i,...,G^{L-1}_i\}$ whose level number and downsample factor are $L$ and $s$ respectively, $O^l$ ($O^l(x,y)\in \{yes, no\}$ represents whether pixel $(x,y)$ is an outlier) is the outlier record at $l$th level, and $F^l$ is the correspondence field at $l$th level. To compute $G^l_i$, we first build the image pyramid $\{I^0_i, I^1_i,...,I^{L-1}_i\}$ with same level number and downsample factor. Then, $G^l_i$ can be obtained by computing the gradient image of $I^l_i$ according to Section~\ref{Sec: Image Gradients}. %\textbf{Compared with , $\{G^0, G^1,...,G^{L-1}\}$ may capture structure information of different scales.}

%After gradient pyramid is known, we can compute the corresponding field with the hierarchical gradient matching framework shown in Figure~\ref{Fig: outline of hierarchical matching}. (as Section~\ref{Sec: gradient matching})
As shown in Figure~\ref{Fig: outline of hierarchical matching}, we first initialize $F^m$ ($m=2$ in this figure) as the KDtree-based initialization step of \cite{NNF_kdtrees}. %and perform our basic gradient matching at $m$th level (see Figure~\ref{Fig: refinement of initial field}(a,b) for an example).
Then, refinement of initial field or propagation of the refined field is a process of iteratively performing following steps:
\begin{enumerate}[\emph{Step~}1.]
\item Basic gradient matching;
\item Forward-backward consistency check;
\item Outlier record $O^l$ update;
\item If required (have blue arrow pointing towards following level), correspondence field propagation;
\item If required (have red arrow pointing towards following level), outlier record propagation.
\end{enumerate}
Note that, as shown in Figure~\ref{Fig: outline of hierarchical matching}, not every iteration requires \emph{Step}~4 or \emph{Step}~5.

\begin{figure*}[t]
\begin{center}
   \includegraphics[width=0.99\linewidth]{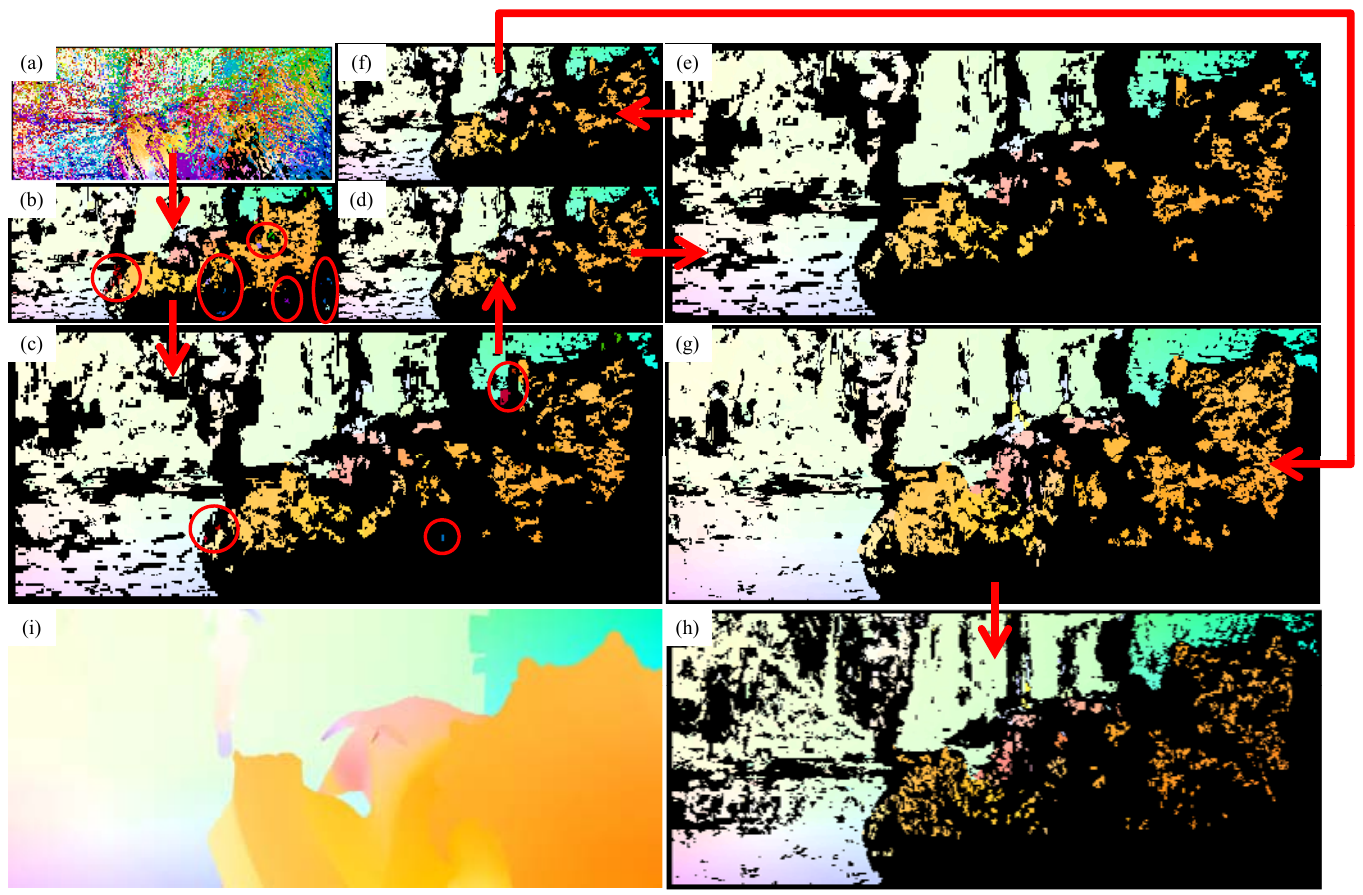}
\end{center}
   \caption{A visual example of correspondence fields generated by our pyramidal gradient matching. (a) is the KDtree-based initial field at 2nd level. (b) is the result of basic gradient matching which takes (a) as initialization. (c), (d), (e) and (f) are fields generated in different iterations of the refinement of initial field. Different from (c-f), from (f) to (g), we only propagate correspondence field. (h) is the final field (displaying at 50\%) of our method. (i) is the ground truth (displaying at 50\%). Along with the increasing of iteration times, we can find that outliers (for example, red circles in (b) and (c)) are removed gradually.}
\label{Fig: refinement of initial field}
\end{figure*}

For the sake of clarity, we give a detailed description of \emph{Step}~3-5. Suppose $C^l$ (similar to $O^l$, $C^l(x,y)\in \{yes, no\}$) is the result of forward-backward consistency check at $l$th level. In \emph{Step}~3, we update $O^l$ as
\begin{equation}
%\label{discriminate moving direction}
\small{
O^l(x,y)=
\left\{
\begin{array}{cc}
O^{l}(x,y) \wedge C^l(x,y),&\text{if}~O^{l}(x,y)~\text{is initialized};\\
C^l(x,y), &\text{else};
\end{array}
\right.
,
}
\end{equation}
where $\wedge$ is logical AND, and ``$O^{l}(x,y)$ is initialized'' means that outlier record propagation is required in \emph{Step}~5 of last iteration. In fact, except the bottom level, we need to propagate correspondence field at all the other levels in \emph{Step}~4. In detail, we propagate field $F^j$ at $j$th level to following $k$th$\in \{j+1,j-1\}$ level as
\begin{equation}
\label{eq: propagate field}
F^k(x,y)=
\left\{
\begin{array}{cc}
uninitialized,&\text{if}~\aleph' = \emptyset;\\%\text{if}~O^{j}(p)= yes~\text{for all}~p\in\aleph
 \frac{\sum \limits_{p\in\aleph'} F^j(p)}{\eta \sum \limits_{p\in\aleph'} 1}, &\text{else};\\
\end{array}
\right.
,
\end{equation}
where
%\begin{equation*}
%\aleph=
%\left\{
%\begin{array}{cc}
%\{(x',y')|0\leq x'-\eta x < \eta,0\leq y'-\eta y < \eta\},&\text{if}~k=j-1;\\
%\{(\eta x,\eta y)\}, &\text{else if}~k=j+1;\\
%\end{array}
%\right.
%,
%\end{equation*}
\begin{multline*}
%\label{discriminate moving direction}
\text{if}~k=j+1\\
\shoveleft{
\left\{
\begin{array}{l}
\eta= \frac{1}{s};\\
\aleph=\{(x',y')|0\leq x'-\eta x < \eta,0\leq y'-\eta y < \eta\};\\
\end{array}
\right.}\\
\shoveleft{\text{else if}~k=j-1}\\
\shoveleft{\left\{
\begin{array}{l}
\eta= s;\\
\aleph=\{(\eta x,\eta y)\};\\
\end{array}
\right.}
\end{multline*}
and $x,y$ start from zero, $\aleph'=\{p\in\aleph|O^{j}(p)= no\}$. Similarly, in \emph{Step}~5, we propagate outlier record $O^j$ at $j$th level to following $k$th$\in \{j+1,j-1\}$ level as
\begin{equation}
O^k(x,y)=
\left\{
\begin{array}{cc}
yes,&\text{if}~\aleph' = \emptyset;\\%\text{if}~O^{j}(p)= yes~\text{for all}~p\in\aleph
no, &\text{else};\\
\end{array}
\right.
,
\end{equation}
where $\aleph'$ is the same as the definition in \emph{Step}~4. A remark is that some pixels of $F^{k}$ are marked as $uninitialized$ (as defined in Formula~(\ref{eq: propagate field})), which means they need different treatments as described in Section~\ref{Sec: gradient matching}.

As shown in Figure~\ref{Fig: refinement of initial field}(a-f), the idea of iteratively propagating correspondence field and outlier record is effective to remove outliers, especially those which may be difficult to be removed just by performing forward-backward consistency check at one level. For example, outliers in regions marked by red circles are well rejected after the first few iterations ``Figure~\ref{Fig: refinement of initial field}: (a)$\rightarrow$(b)$\rightarrow$(c)'', and no outlier is introduced in the remaining iterations.
%except filtering outliers in $F^{m-1}$(or $F^m$) via forward-backward consistency check (see Section~\ref{Sec: Outlier Filtering in One Level}), we also filter outliers through checking between levels. Specially, each pixel of $F^{m-1}$(or $F^m$) is regarded as an inlier only when both itself and its corresponding pixel in $F^m$(or $F^{m-1}$) are consistent.

After $N$ times of refinement of initial field between $m$th level and $(m-1)$th level, like Figure~\ref{Fig: refinement of initial field}(f), we obtain a refined field $F^{m}$ which is more accurate and contains less outliers. Then, we only propagate this refined field to $(m-1)$th level (the single blue arrow in Figure~\ref{Fig: outline of hierarchical matching}). Note that, we do not propagate outlier record here as $F^{m}$ may be too sparse. % after $N$ iterations of outlier propagation in the refinement step, therefore, propagating outlier continually may lead.
%So each pixel of $F^{m-1}$ is regarded as inlier if it passes the forward-backward consistency check at $(m-1)$th level.
However, we still propagate outlier record in the remaining propagation steps from $(m-1)$th level to the full resolution level. As shown in Figure~\ref{Fig: refinement of initial field}(f-h), our method can grow inliers and avoid introducing outliers effectively based the refined initial field.

Compared to the classical coarse-to-fine framework, instead of mainly handling large displacements, our pyramidal matching focuses on robust matching of more challenging cases such as scale/rotational deformations, illumination changes. %In the view of patch matching approaches, pure large displacement is not so challenging.
Compared to FlowFields \cite{FlowFields}, we share the idea of avoiding finding outliers by limited random search. However, the design of our pyramidal matching is quite different (detailed in Section~\ref{Sec: Secret of Limited PatchMatch}), which is tailored to gradient matching. %Especially, based on our analysis that limited PatchMatch is effective to generate continuous field from sparse initialization and has the ability to avoid introducing new outliers which is nonexistent in the initialization, we introduce iterative refinement of initial correspondence field and some special outlier filtering strategies aimed at both inter-level and intra-level (see Section~\ref{Sec: Outlier Filtering}).

\subsection{Outlier Filtering}
\label{Sec: Outlier Filtering}
%\label{Sec: Outlier Filtering among Levels}

\begin{table}[t]
\begin{center}
\caption{The settings of two gradient pyramids used in our approach. Their differences lie in color space (RGB, CIELab and YCrCb). As described in Section~\ref{Sec: Outlier Filtering}, these two gradient pyramids are used to compute bidirectional correspondence fields respectively.}
\label{Tab: settings of gradient pyramid}
\small{
\begin{tabular}{c|c|c|c}
\hline
Level
&0 & 1 & 2 \\
\hline
%\multicolumn{5}{|c|}{Forward gradient pyramid}\\
\hline
Forward gradient pyramid & RGB & CIELab & YCrCb\\
%\hline
%\multicolumn{5}{|c|}{Backward gradient pyramid \Rmnum{1}}\\
\hline
Backward gradient pyramid & CIELab & YCrCb & RGB\\
\hline
\end{tabular}
}
\end{center}
\end{table}

The basic operation of outlier filtering is to compute bidirectional correspondence fields and perform forward-backward consistency check between them. We suppose that outliers are mainly due to accidental factors and belong to unstable local optimal flow, so that their probability of occurring consistently with different settings in hierarchical matching process is low. Under this assumption, we can further enhance outlier filtering by introducing following differences in the matching process.

An effective extension of forward-backward consistency check is to perform two way consistency check \cite{FlowFields}, where two backward correspondence fields are computed with different patch radiuses and each pixel of the forward correspondence field is regarded as an inlier only when it is consistent to both backward correspondence fields. %As mentioned in Section~\ref{Sec: Outlier Filtering among Levels}, we can regard the extension of \cite{FlowFields} as introducing differences in patch size between two matching process.
In our method, we do not adopt two way consistency check, but use two different patch radiuses to compute the forward field ($r=7$) and the backward field ($r=5$).

In cooperation with correspondence field propagation and outlier record propagation in our pyramidal matching, differences among levels and forward/backward fields can serve as multiple outlier sieves. %where matches that arrive one level must pass all checks of its previous levels which are different as these inter-level differences.
A natural inter-level difference in pyramidal matching is that different levels have different gradient image scales and equivalent %(relative to the full resolution image)
patch radiuses. %For example, an outlier may be mistaken for a good match at a level, however, it is likely to be rejected on next level as differences in gradient image scale or equivalent patch radius.
In addition, another difference is that %(1) We adopt different Sobel kernels (identified by kernel size $S$) to compute gradient images; (2)
we convert source images into different color spaces at different levels before computing gradient images of them. Specially, Table~\ref{Tab: settings of gradient pyramid} gives color space settings of two gradient pyramids ($L=3$) used in our approach. %We adopt three different color spaces: RGB, CIELab and YCrCb.
Note that, there are two kinds of differences in Table~\ref{Tab: settings of gradient pyramid}. 1) Inter-level differences: for each row (corresponding to one pyramid), its any two levels have different color spaces (choosing two from RGB, CIELab and YCrCb). 2) Intra-level differences: for each column (corresponding to forward/backward fields of one level), color space between two gradient pyramids are different, for example, the intra-level difference of 1st level lies in CIELab $\leftrightarrow$ YCrCb.

After forward-backward consistency check, the resulting field may still contain some small regions. %We suppose that their possibility of being outliers is very high, therefore,
As \cite{FlowFields}, we segment the field and then remove regions whose area are too small ($<9$ pixels in our experiments) in the segmentation result.% In the segmentation of correspondence field, we mark neighboring pixels with same label if the difference between their motion is small enough ($<12$ pixels in our experiments).% 

\subsection{Scalability of Gradient Image}
\label{Sec: Flexibility of Gradients}

In this section, we will discuss the scalability of gradient image. Here, ``scalability'' means that we can use different levels of gradient feature to obtain different complexity without dramatic accuracy change, which also demonstrates the robustness of gradient image for optical flow estimation. In this paper, we present four versions with different levels of gradient image:
\begin{enumerate}[(1)]
\item PGM-C: The full version of our approach. The gradient image is 6-channel and computed from color image according to Section~\ref{Sec: Image Gradients}.
\item PGM-G: The difference between PGM-G and PGM-C is that we replace color image with gray image, thus, the gradient image is 2-channel in PGM-G.
\item PGM-CD: Compared to PGM-C, PGM-CD only uses the direction information of gradient image for patch matching. Specially, the gradient image used in PGM-CD is converted from 6-channel gradient image $G$ as
    $$G(x,y)=G(x,y)./|G(x,y)|,$$
    where $./$ is to perform array division by dividing each element of its left operand by the corresponding element of its right operand, and $|G(x,y)|$ is to compute the absolute value of each element of $G(x,y)$.
\item PGM-GD: Similar to PGM-CD, PGM-GD only uses the direction information of the gradient image used PGM-G.
\end{enumerate}

Through choosing different levels of gradient image, we can change the complexity of gradient matching. More important, as shown in Section~\ref{Sec: Effectiveness and scalability}, their accuracy will not change abruptly.

\subsection{Sparse-to-Dense Interpolation}
\label{Sec: Sparsification and Dense}

After obtaining the final correspondence field at the full resolution level, we select pixels which are on the cross points of the regular grid (grid spacing is 3 pixels) from the final correspondence field and pass these matches to EpicFlow \cite{EpicFlow} to generate dense optical flow field. Except interpolator parameter (NW or LA), we use the online code with default parameters\footnote{All our experiments use the executable file \emph{epicflow-static} provided by \url{http://lear.inrialpes.fr/src/epicflow/} as: \emph{epicflow-static img1 img2 edge matches flo -sintel/kitti/middlebury -nw}, where \emph{-nw} is added selectively.} for three datasets unless particularly stated. In our method, we choose LA as our interpolator if enough matches ($>2.2\%$ of the image size) are input into EpicFlow, otherwise NW is chosen. Through this strategy, we can avoid magnifying error when matches are too sparse or unreliable for local affine transformation which is simply identified by the number of matches.

%\begin{equation}
%%\label{discriminate moving direction}
%Interpolator=
%\left\{
%\begin{array}{cc}
%NW,&if~N_{matches}<N_T;\\
%LA, &other;
%\end{array}
%\right.
%,
%\end{equation}

\section{Experiments}
\label{Sec: Experiments}
In this section, we evaluate our method on three optical flow datasets:
\\$\bullet$ MPI Sintel dataset \cite{MPI-benchmark} is a challenging benchmark based on an animated short film. It features long sequence, large displacements, motion/defocus blur, occlusions, atmospheric effects and so on. Two kinds of passes (clean pass and final pass) are available.
\\$\bullet$ KITTI dataset \cite{KITTI-benchmark,KITTI2015} consists of images of city streets which are captured by a camera fixed on a driving platform. Most motions of KITTI dataset are caused by camera translation and rotation. Note that, in addition to KITTI flow 2012, KITTI flow 2015 is released recently, and we call them KITTI2012 and KITTI2015 respectively.
\\$\bullet$ Middlebury dataset \cite{middlebury-benchmark} is a classical dataset used for evaluating optical flow algorithms. Its test set consists of 12 image pairs which contain complex motions, but displacements are limited.

Our experiments are organized as follows. Firstly, to demonstrate the effectiveness of our gradient image and pyramidal matching framework, we evaluate the four versions (described in Section~\ref{Sec: Flexibility of Gradients}) on training sets of MPI Sintel and KITTI2012 datasets in Section~\ref{Sec: Effectiveness and scalability}. Then, we generate optical flow results of our method on test sets of three datasets and compare to state of the art methods with same parameters (see Section~\ref{Sec: MPI}, Section~\ref{Sec: KITTI} and Section~\ref{Sec: Middlebury}).

We optimize the parameters of PGM-C on a subset ($20\%$) of the MPI Sintel training set and do not change them for other three versions (PGM-G, PGM-CD and PGM-GD) and different datasets, which demonstrates the robustness of our method. Specially, we use \{$W=2$, $S=5$, $L=3$, $s=\frac{1}{2}$, $N=2$, $m=2$\} and perform 6 times of propagation-search iteration in basic gradient matching at full resolution level and 4 times at other levels.  A remark is that some settings (like using different color space as Table~\ref{Tab: settings of gradient pyramid}) are meaningless for PGM-G and PGM-GD. All experiments are conducted on a PC (Intel Pentium G3240 CPU @3.1GHz and 4GB Memory), and no parallel techniques (such as GPU, OpenMP, SSE) are used.

\subsection{Effectiveness of our features and matching framework}
\label{Sec: Effectiveness and scalability}

In order to demonstrate the effectiveness of our gradient image and pyramidal matching framework, we evaluate our method on training sets of MPI Sintel and KITTI2012 datasets. As a result of the scalability of gradient image, four versions of our method are available. In addition, the results of the online code\footnote{We generate results of CPM-Flow by using the executable file \emph{cpm.exe} provided by \url{http://42.96.142.148/owncloud/s/xu6Q8d9ZdVFIDre} as: \emph{cpm.exe img1 img2 matches 1(or 3) 4 5 6}.} of CPM-Flow \cite{CPMFlow}, which is more efficient than state of the art methods with similar quality, are also given for comparison. Note that, we use EpicFlow with same parameters for all rows of Table~\ref{Tab: evaluation of effectiveness and scalability} in this section.

\begin{table}[t]
\begin{center}
\caption{Comparison of CPM-Flow and different versions of our method. We test different versions and methods on training sets of MPI Sintel (2082 image pairs in total) and KITTI2012 (194 image pairs in total). AEE on each pass is computed with all 1041 image pairs of corresponding pass. Out-Noc (ALL) is the percentage of erroneous pixels ($>3$) in non-occluded (all) areas. Column ``Time'' is the average time required for processing corresponding dataset under the same running environment, and interpolation time is not included. Note that, all results in this table are obtained by using EpicFlow with same parameters.}
\label{Tab: evaluation of effectiveness and scalability}
{
\begin{tabular}{|@{}C{58pt}@{}|@{}C{42pt}@{}|@{}C{42pt}@{}|@{}C{22pt}@{}||@{}C{28pt}@{}|@{}C{28pt}@{}|@{}p{22pt}@{}|}
\hline
\multirow{3}{*}{Method}&\multicolumn{3}{c||}{MPI Sintel} & \multicolumn{3}{c|}{KITTI2012}\\
\cline{2-7}
& \mytabincell{c}{AEE\\(Clean Pass)}  & \mytabincell{c}{AEE\\(Final Pass)} & Time & \mytabincell{c}{Out-\\Noc} & \mytabincell{c}{Out-\\All} & ~Time \\
\hline
FF\cite{FlowFields}+Gradients& 2.875 & 3.997  & --  &  15.31\% &  23.68\% & ~~~--\\
PGM+CIELab& 2.405 & 5.276  & --  &  13.40\% &  21.68\% & ~~~--\\
PGM+SIFT\cite{SIFTFlow}& 2.029 &  3.520 & --  &  6.68\% &  15.46\% & ~~~--\\
\hline
PGM-C& 1.881 & \textbf{3.407}  & 3.49s  &  5.58\% &  14.24\% & ~4.55s\\
\hline
PGM-G& 1.910 & 3.541  & 1.82s  & \textbf{5.14\%}  & \textbf{13.79\%}  & ~2.05s\\
PGM-CD&1.936 & 3.674  & 1.56s  & 6.16\%  & 14.95\% & ~1.60s\\
PGM-GD&2.104 & 4.024  & \textbf{1.08s}  & 5.58\%  & 14.32\% & ~\textbf{1.16s}\\
\hline
PGM-C3&\textbf{1.872} & 3.414  &  2.28s & 5.53\% & 14.24\% & ~2.77s\\
PGM-G3&1.986 & 3.647  & 1.13s  & 5.23\%  & 13.92\% & ~1.26s\\
%PGM-Cp&2.126 & 3.792  &   & \%  & \% &\\
\hline
CPM-d3\cite{CPMFlow}&2.136& 3.633  & 2.41s  & 6.18\%  & 14.80\% & ~2.49s\\
CPM-d1\cite{CPMFlow}&1.991& 3.599  & 10.61s & 6.48\%  & 15.10\% & 10.81s\\
\hline
\end{tabular}
}
\end{center}
\end{table}

\begin{figure*}[t]
\begin{center}
\includegraphics[width=0.99\linewidth]{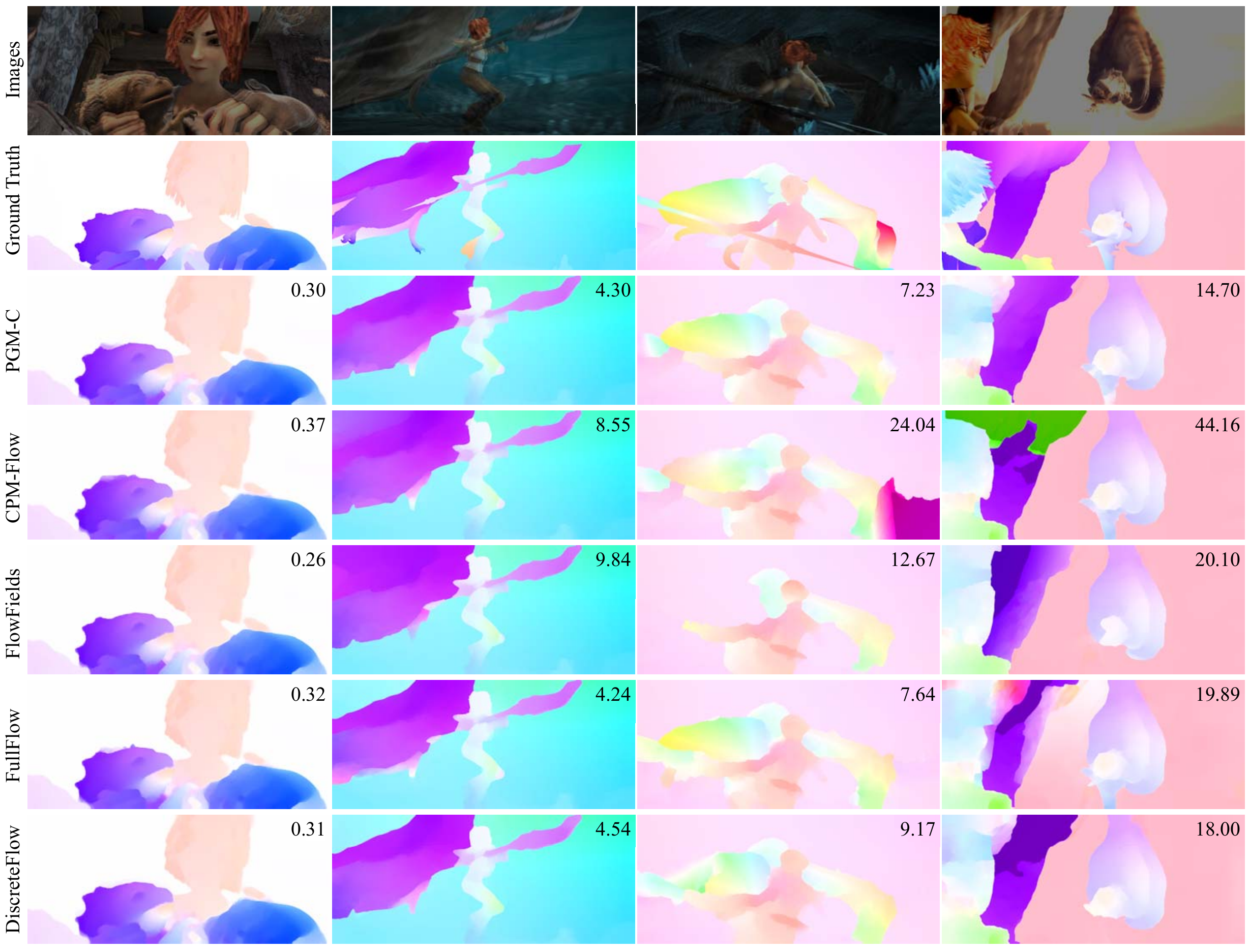}
\end{center}
\caption{Visual examples on MPI Sintel benchmark. From top to bottom, each row shows: average image of two consecutive images, ground truth, results of our PGM-C and results of 4 state of the art methods (CPM-Flow \cite{CPMFlow}, FlowFields \cite{FlowFields}, FullFlow \cite{FullFlow} and DiscreteFlow \cite{DiscreteFlow}). From left to right, the first column is a simple case with small motion, and the remaining are three challenging cases (like large displacements, scale/rotational deformations). AEE is given on top right corner.}
\label{Fig: Visual examples on MPI}
\end{figure*}

\begin{figure}[t]
\begin{center}
   %\includegraphics[width=0.4\linewidth]{grts_channel1.pdf}
   \includegraphics[width=0.99\linewidth]{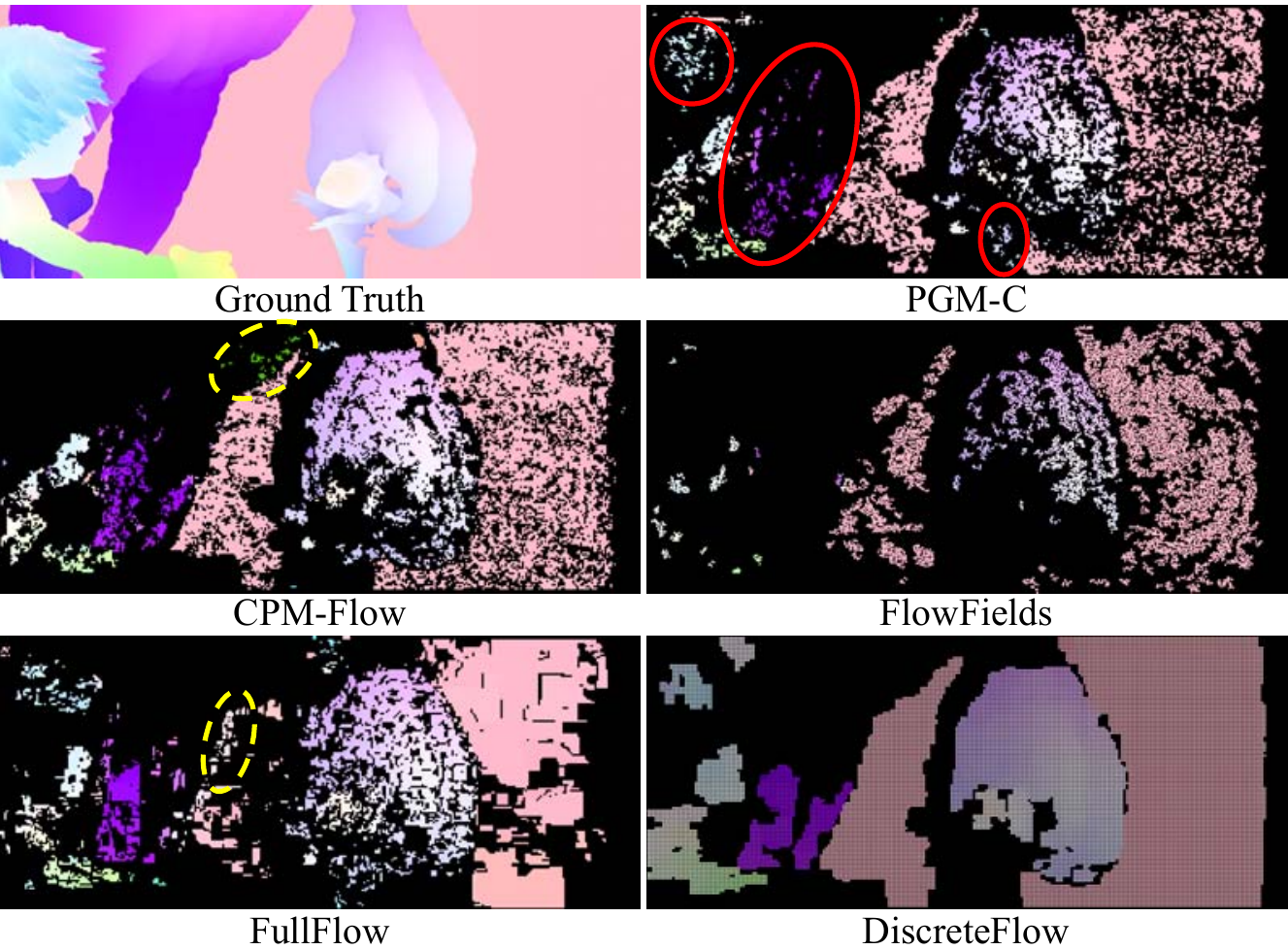}
\end{center}
   \caption{An visual example of matches produced by five methods. These matches correspond to the fourth column of Figure~\ref{Fig: Visual examples on MPI}. Note: these figures are drawn directly from matches which are passed to EpicFlow, and each match is drawn as a $3 \times 3$ filled rectangle for clear visual effects.}
\label{Fig: Example of matches on PMI}
\end{figure}

It is indeed incredible that our simple gradient image---one of main contributions ---is effective for highly accurate optical flow estimation. In order to clear up doubt and demonstrate the effectiveness of both our gradient image and our pyramidal matching framework, especially compared to FlowFields\cite{FlowFields}, we first test three combinations: \begin{enumerate}[(1)]
\item FF\cite{FlowFields}+Gradients: Based on FlowFields, but replace SIFT features with our gradient image,
\item PGM+CIELab: Based on our PGM-C, but replace gradient image with CIELab color image,
\item PGM+SIFT\cite{SIFTFlow}: Based on our PGM-C, but replace gradient image with SIFT features\cite{SIFTFlow}.
\end{enumerate}
The performance comparison of FF\cite{FlowFields}+Gradients, PGM+CIELab, PGM+SIFT\cite{SIFTFlow} and PGM-C is given in the 1st$\sim$4th row of Table~\ref{Tab: evaluation of effectiveness and scalability}. We can find that the error of PGM-C is the least. Especially, the error of PGM-C is far less than FF\cite{FlowFields}+Gradients and PGM+CIELab, for example, Out-Noc of PGM-C on KITTI2012 is $<0.5$ times of these two. The comparison among PGM-C, PGM+CIELab and PGM+SIFT\cite{SIFTFlow} demonstrates the effectiveness of our gradient image, and the comparison between PGM-C and FF\cite{FlowFields}+Gradients demonstrates the effectiveness of our pyramidal matching framework. So the high accuracy of our method depends on both gradient image and pyramidal matching framework. Compared to the hierarchical framework of FlowFields, our pyramidal matching framework is more efficient and better for gradient images. Compared to SIFT features and CIELab color image, our gradient image is simple and more suitable for pyramidal matching framework.

The remaining rows of Table~\ref{Tab: evaluation of effectiveness and scalability} compare the results of our method and CPM-Flow. For our method, except four versions (PGM-C, PGM-G, PGM-CD and PGM-GD) described in Section~\ref{Sec: Flexibility of Gradients}, two derivative versions (PGM-C3 and PGM-G3) of PGM-C and PGM-G are also tested. In PGM-C3 and PGM-G3, similar to CPM-Flow, we reduce computational complexity by ignoring random search steps at the full resolution level of pixels who are not on the cross points of the regular grid (as Section~\ref{Sec: Sparsification and Dense}, grid spacing is 3 pixels). For CPM-Flow, we report its performance with different grid spacing $d$ (see \cite{CPMFlow}): CPM-d3 means $d=3$ and CPM-d1 means $d=1$.

As shown in Table~\ref{Tab: evaluation of effectiveness and scalability}, we can find that AEEs of PGM-C, PGM-G, PGM-CD and PGM-GD on MPI Sintel dataset increase gradually. However, their accuracy is controllable, which demonstrates the scalability and robustness of gradient image. For example, the ratio between standard deviation and mean value of AEEs (the second and third column of Table~\ref{Tab: evaluation of effectiveness and scalability}) of our four versions on two passes of MPI Sintel training set are 5.11\% and 7.24\% respectively, %, 7.45\%, 3.33\%.
while the ratio of the top 10 methods in Table~\ref{Tab: Performance of PGM-C on MPI} (including our PGM-C) on two passes of MPI Sintel test set are 15.81\% and 9.20\% respectively. It shows that the accuracy changes of our four versions are controllable. As final pass contains more challenging cases (such as motion/defocus blurs and atmospheric effects) than clean pass, performance fluctuation of the four versions on final pass is more significant.
For KITTI2012 dataset, PGM-G is even more accurate than PGM-C with fast speed.

In aspect of speed, the time cost decreases progressively from PGM-C to PGM-GD. Therefore, our method is able to provide a good compromise between speed and accuracy. Especially, compared to PGM-C, PGM-G achieves significant speed-up with little accuracy loss on both benchmarks. The speed advantage of our method is obtained by our gradient image. Although we compute $L_2$ distance between two patches as Eq.~\ref{Eq: Define of patch distance} (CPM-Flow computes $L_1$ distance) and our pyramidal matching framework iteratively performs basic gradient matching which introduce extra computation, gradient image is efficient enough to support fast matching at different levels. For example, in terms of feature dimension, gradient image feature of each pixel in PGM-C is 6 dimensions, while the SIFT flow feature used in \cite{CPMFlow} is 128 dimensions.

\begin{table}[t]
\begin{center}
\caption{Performance comparison of top 10 published methods on MPI Sintel test set. This table reports AEE in three kinds of areas: all areas (AEE-All), non-occluded areas (AEE-Noc) and occluded areas (AEE-Occ). Bold results represent the best and underlined results are the second best.}
\label{Tab: Performance of PGM-C on MPI}
\small{
\begin{tabular}{|@{}C{82pt}@{}|@{}C{32pt}@{}|@{}C{32pt}@{}|@{}C{32pt}@{}|c|}
\hline
\mytabincell{c}{Method\\(Clean Pass)}
 & \mytabincell{c}{AEE-\\All} & \mytabincell{c}{AEE-\\Noc} & \mytabincell{c}{AEE-\\Occ} & Time\\
\hline
\textbf{PGM-C}                      & \textbf{3.234}	& \textbf{0.929}	& \textbf{22.045}  & \underline{3.49s+3s}\\
CPM-Flow\cite{CPMFlow}              & \underline{3.557}	& 1.189	& 22.889 & \textbf{4.3s}\\
DiscreteFlow\cite{DiscreteFlow}     & 3.567	& 1.108	& 23.626 & $\sim$180s\\
FullFlow\cite{FullFlow}             & 3.601	& 1.296	& \underline{22.424} & $\sim$84s\\
FlowFields\cite{FlowFields}         & 3.748	& \underline{1.056}	& 25.700 & 18s\\
EpicFlow\cite{EpicFlow}             & 4.115	& 1.360	& 26.595 & 16.4s\\
PH-Flow\cite{PHflow}                & 4.388	& 1.714	& 26.202 & $>$200s\\
AggregFlow\cite{Aggregation}        & 4.754	& 1.694	& 29.685 & $>$1620s\\
TF+OFM\cite{Kennedy2015}            & 4.917	& 1.874	& 29.735 & 500s\\
Deep+R\cite{DB15c}                  & 5.041	& 1.481	& 34.047 & 179s\\
%DeepFlow2                           & 4.891	& 1.403	& 33.317 & \\
\hline
\multicolumn{5}{c}{}\\
\hline
\mytabincell{c}{Method\\(Final Pass)}
 & \mytabincell{c}{AEE-\\All} & \mytabincell{c}{AEE-\\Noc} & \mytabincell{c}{AEE-\\Occ} & Time\\
\hline
\textbf{PGM-C}                      & \textbf{5.591} & \underline{2.672} & \textbf{29.389}  & \underline{3.49s+3s}\\
FlowFields\cite{FlowFields}         & \underline{5.810}& \textbf{2.621}& 31.799 & 18s\\
FullFlow\cite{FullFlow}             & 5.895	& 2.838	& 30.793  & $\sim$84s\\
CPM-Flow\cite{CPMFlow}              & 5.960	& 2.990	& \underline{30.177} & \textbf{4.3s}\\
DiscreteFlow\cite{DiscreteFlow}     & 6.077	& 2.937	& 31.685 & $\sim$180s\\
EpicFlow\cite{EpicFlow}             & 6.285	& 3.060	& 32.564 & 16.4s\\
TF+OFM\cite{Kennedy2015}            & 6.727	& 3.388	& 33.929 & 500s\\
Deep+R\cite{DB15c}                  & 6.769	& 2.996	& 37.494 & 179s\\
SparseFlowFused\cite{SparseFlow}    & 7.189	& 3.286	& 38.977 & $\sim$10s\\
DeepFlow\cite{deepflow}             & 7.212	& 3.336	& 38.781 & 19s\\
\hline
\end{tabular}
}
\end{center}
\end{table}

Compared to CPM-d3, most of our four versions have higher accuracy. For example, all four versions have less AEE than CPM-d3 on clean pass of MPI Sintel training set, and less Out-Noc on KITTI2012 training set. Except PGM-C, other three versions are faster than CPM-d3. CPM-d1 has a better performance than CPM-d3, however, it is significantly slower ($>10$s). The high efficiency of CPM-d3 is brought by only finding matches of some seeds rather than every pixel of the image, which may also be effective for our method. In order to verify its effectiveness on our method, we also compare the results of PGM-C3 and PGM-G3 in Table~\ref{Tab: evaluation of effectiveness and scalability}. These results show that PGM-C3 (or PGM-G3) achieves similar accuracy with faster speed than PGM-C (or PGM-G), and PGM-C3 even has less AEE than PGM-C on clean pass of MPI Sintel training set. It suggests that ignoring some pixels is also effective for our method. A remark is that PGM-C3 and PGM-G3 are a simple verification where we only ignore some random search steps at the full resolution level, while CPM-Flow ignores both random search step and propagation step of some pixels at all levels.

\subsection{Performance on MPI Sintel Benchmark}
\label{Sec: MPI}

Our results compared to other top 9 published methods on two passes of MPI Sintel test set are given in Table~\ref{Tab: Performance of PGM-C on MPI} (captured from MPI Sintel Website\footnote{\url{http://sintel.is.tue.mpg.de/}} on February 18th, 2017). It shows that our PGM-C currently ranks 1st on clean pass and final pass among all published methods, and ranks 3rd on clean pass and 6th on final pass among all submitted methods (taking into account unpublished methods).

As shown in Table~\ref{Tab: Performance of PGM-C on MPI}, our method clearly outperforms most published methods on both clean pass and final pass%, and we can reduce the AEE over the complete frames by 0.323 on clean pass and 0.219 on final pass
. Our method also has a competitive speed, which is just a slightly slower than CPM-Flow (Note: the time reported in Table~\ref{Tab: Performance of PGM-C on MPI} is not normalized for programming environment, cpu speed, number of cores and so on). Compared to other 9 methods, an advantage of our method is that it obtains stable performance (close ranking) on both passes, which demonstrates the robustness of our method. Especially, except non-occluded areas of final pass, our method reduces AEE significantly for other cases.

Figure~\ref{Fig: Visual examples on MPI} gives a visual comparison of four state of the art methods on four image pairs which consist of one general case and three challenging cases. AEEs (the top right corner) show that our method is more robust for these four pairs. For general case, our method has equal capacity to handle small motion, thin objects and motion edges (see the first column of Figure~\ref{Fig: Visual examples on MPI}, especially the eye, mouth and claw of the monster). For some challenging cases, our method can not only find more inliers (such as the top left area of the second column, the head of the monster in the third column, the tail of the big monster and the wing of the small monster in the fourth column), but also effectively reject outliers (for example, the bottom right area of the third column). An example of matches (corresponding to the fourth column of Figure~\ref{Fig: Visual examples on MPI}) of five methods is shown in Figure~\ref{Fig: Example of matches on PMI}. In these qualitative results, our method is able to produce more accurate and dense matches. Note that, compared to other methods, our method especially can better capture motions in all three challenging areas which are marked by red circles in Figure~\ref{Fig: Example of matches on PMI} (the head of the character, the tail of the big monster and the wing of the small monster). For example, CPM-Flow and FlowFields miss the motion of the head, FlowFields and DiscreteFlow miss the motion of the wing, only the bottom half of the tail is captured in FulFlow and DiscreteFlow, and FlowFields only captures a very small area of the tail. In addition, our method also avoids finding serious outliers, such as the areas marked by yellow dotted circles in Figure~\ref{Fig: Example of matches on PMI} (see the fourth column of Figure~\ref{Fig: Visual examples on MPI} for their final interpolation flow).

\subsection{Performance on KITTI Benchmark}
\label{Sec: KITTI}

\begin{figure*}[t]
\begin{center}
\includegraphics[width=0.99\linewidth]{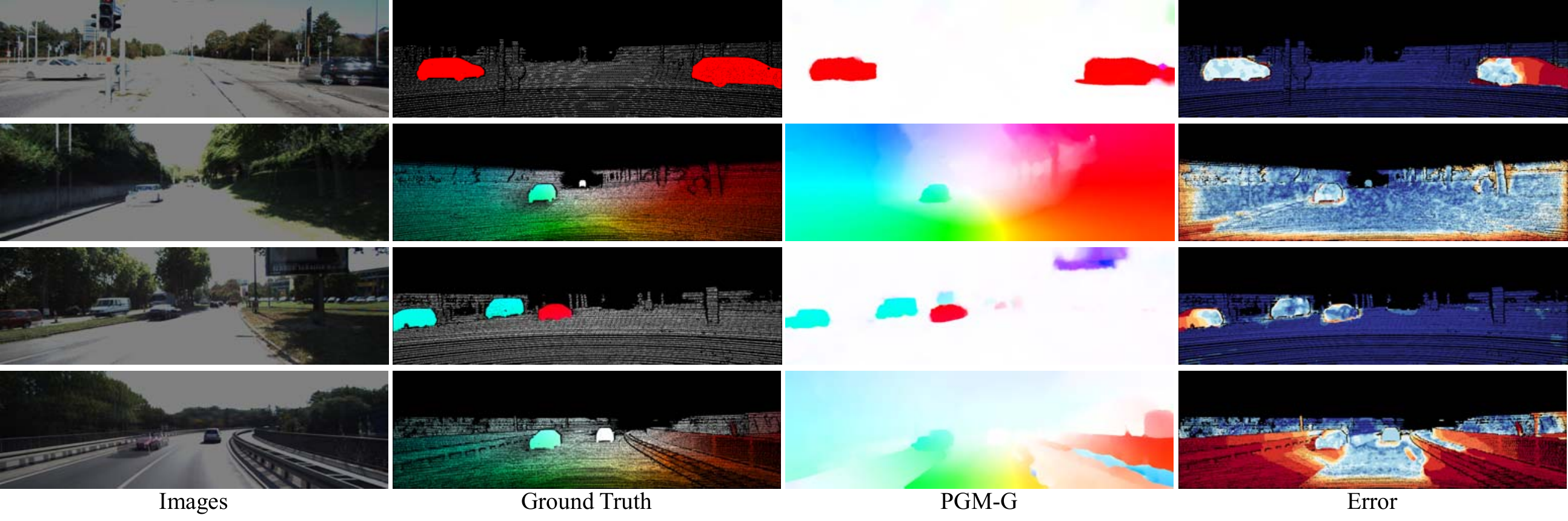}
\end{center}
\caption{Visual examples on KITTI benchmark. From left to right, each column shows: average image of two consecutive images, ground truth, results of our PGM-C and visual error image. Note that, the fourth row is a failure example of our method.}
\label{Fig: Visual examples on KITTI}
\end{figure*}

As captured from a driving platform, KITTI datset contains large displacements, complex lighting conditions and strong scale deformations (caused by high view angle and particularly serious in boundary region). Therefore, general optical flow methods seem to perform not well. Most of the top-ranked methods use special techniques like epipolar geometry and stereo vision.

\begin{table}[t]
\begin{center}
\caption{Performance comparison on KITTI2012 test set. Methods that use epipolar geometry or stereo vison are not included in this table. Runtime marked by $^{*}$ is measured on GPU.}
\label{Tab: Performance of PGM-G on KITTI}
\small{
\begin{tabular}{|@{}C{78pt}@{}|@{}C{32pt}@{}|@{}C{32pt}@{}|@{}C{28pt}@{}|@{}C{28pt}@{}|c|}
\hline
Method
 & \mytabincell{c}{Out-\\Noc} & \mytabincell{c}{Out-\\All} & \mytabincell{c}{AEE-\\Noc} & \mytabincell{c}{AEE-\\All} & Time\\
\hline
PatchBatch\cite{PatchBatch} & \textbf{5.29\%}	& 14.17\%	 & \underline{1.3}	& 3.3 & 50s$^{*}$ \\
\textbf{PGM-G}   & \underline{5.59\%}	& 13.87\%	 & \textbf{1.2}	& 3.3 & \underline{2.05s+3s}\\
PH-Flow\cite{PHflow}& 5.76\% & \textbf{10.57\%} & \underline{1.3} & \textbf{2.9} & 800s\\
FlowFields\cite{FlowFields}& 5.77\% & 14.01\% & 1.4 & 3.5 & 23s\\
CPM-Flow\cite{CPMFlow}  & 5.79\% & 13.70\% & \underline{1.3} & \underline{3.2} & \textbf{4.2s}\\
NLTGV-SC\cite{NLTGV}  & 5.93\%  & \underline{11.96\%}  & 1.6  & 3.8  & 16s$^{*}$\\
TGV2ADCSIFT\cite{TGV2ADCSIFT}      &6.20\%	&15.15\%	& 1.5	& 4.5 & 12s$^{*}$\\
DiscreteFlow\cite{DiscreteFlow} & 6.23\% & 16.63\% & \underline{1.3} & 3.6 & 180s\\
DeepFlow\cite{deepflow} & 7.22\% & 17.79\% & 1.5 & 5.8 & 17s\\
EpicFlow\cite{EpicFlow} & 7.88\% & 17.08\% & 1.5 & 3.8 & 15s\\
\hline
\end{tabular}
}
\end{center}
\end{table}

\begin{table}[t]
\begin{center}
\caption{Performance comparison on KITTI2015 test set. Fl-Bg (Fg) is
the percentage of outliers averaged only over background (foreground) regions. As Table~\ref{Tab: Performance of PGM-G on KITTI}, we exclude non optical flow methods and mark GPU methods by $^{*}$.}
\label{Tab: Performance of PGM-G on KITTI2015}
\small{
\begin{tabular}{|@{}C{82pt}@{}|@{}C{38pt}@{}|@{}C{38pt}@{}|@{}C{38pt}@{}|c|}
\hline
Method
 & Fl-Bg & Fl-Fg & Fl-All & Time\\
\hline
\textbf{PGM-G}   & \textbf{18.90}\% & 	28.21\% & 	\textbf{20.45}\% & \underline{2.05s+3s}\\
PatchBatch\cite{PatchBatch} & \underline{19.98\%}	& 30.24\% &	\underline{21.69\%} & 50s$^{*}$ \\
DiscreteFlow\cite{DiscreteFlow} & 21.53\%   & \textbf{26.68}\%	&22.38\% & 180s\\
CPM-Flow\cite{CPMFlow}  & 22.32\%	&\underline{27.79\%}	&23.23\% & \textbf{4.2s}\\
FullFlow\cite{FullFlow} & 23.09\%	&30.11\%	&24.26\% & 240s\\
EpicFlow\cite{EpicFlow} & 25.81\%   &33.56\%	&27.10\% & 15s\\
DeepFlow\cite{deepflow} & 27.96\%	&35.28\%	&29.18\% & 17s\\
\hline
\end{tabular}
}
\end{center}
\end{table}

According to Table~\ref{Tab: evaluation of effectiveness and scalability}, we evaluate PGM-G on KITTI dataset. Note that, except differences described in Section~\ref{Sec: Flexibility of Gradients}, all other settings are the same for four versions and different datasets. In addition to KITTI2012, we also test our method on the recently released KITTI2015. Table~\ref{Tab: Performance of PGM-G on KITTI} and Table~\ref{Tab: Performance of PGM-G on KITTI2015} report the results (captured on February 18th, 2017) on KITTI2012\footnote{\url{http://www.cvlibs.net/datasets/kitti/eval_stereo_flow.php?benchmark=flow}} and KITTI2015\footnote{\url{http://www.cvlibs.net/datasets/kitti/eval_scene_flow.php?benchmark=flow}} for our method and state of the art published methods which do not use special techniques such as epipolar geometry, stereo vision. We can see that PGM-G ranks 1st on KITTI2015 and 2nd on KITTI2012. Our method clearly outperforms EpicFlow, FlowFields, CPM-Flow and DiscreteFlow on both KITTI2012 and KITTI2015. On KITTI2012, %PGM-G missed the best approach by 0.3\% in Out-Noc, but
PGM-G has smallest AEE-Noc. On KITTI2015, our method has least Fl-Bg and Fl-All. PGM-G also has a significant advantage on speed, which is faster than CPM-Flow under the same running environment (see Table~\ref{Tab: evaluation of effectiveness and scalability}). Qualitative results on KITTI are given in Figure~\ref{Fig: Visual examples on KITTI}. As shown in the first three rows, our method is able to capture the motion of cars and background well. The fourth row shows a failure case of our method due to containing large homogeneous regions (like the road surface). In homogeneous regions, patch matching is easy to be affected by noise or disturbance and find outliers.

\subsection{Performance on Middlebury Benchmark}
\label{Sec: Middlebury}

\begin{table}[t]
\begin{center}
\caption{Performance comparison on Middlebury test set. This table compares four methods which are based on sparse-to-dense interpolation.}
\label{Tab: Performance of PGM-C on Middlebury}
\small{
\begin{tabular}{|@{}C{62pt}@{}|@{}C{42pt}@{}|@{}C{42pt}@{}||@{}C{42pt}@{}|@{}p{43pt}@{}|}
\hline
Method
 & Avg. AEE & AEE rank & Avg. AAE & ~AAE rank\\
\hline
FlowFields\cite{FlowFields} & \textbf{0.331}	& \textbf{40.1}	&\textbf{3.175} & ~~~~\textbf{47.1}\\
\textbf{PGM-C}              & \underline{0.349} & \underline{48.9}  &\underline{3.303} & ~~~~\underline{48.9}\\
CPM-Flow\cite{CPMFlow}      & 0.368	& 52.1	&3.423 & ~~~~53.9\\
EpicFlow\cite{EpicFlow}     & 0.393 & 55.6	&3.545 & ~~~~54.2\\
\hline
\end{tabular}
}
\end{center}
\end{table}

As discussed in \cite{EpicFlow,FlowFields,CPMFlow}, the benefits that can be obtained by matching+EpicFlow on Middlebury are limited because of the lack of large displacements. We test PGM-C on Middlebury dataset without setting changes. Our PGM-C obtains an average AEE of 0.349 and an average AAE of 3.303. A performance comparison of some sparse-to-dense interpolation based methods on Middlebury dataset is given in Table~\ref{Tab: Performance of PGM-C on Middlebury}, and our method performs slightly better than CPM-Flow and EpicFlow.

\section{Conclusions}
\label{Conclusions Section}

In this paper, we propose a pyramidal gradient matching approach that can provide dense matches for highly accurate and efficient optical flow estimation. A novel contribution in this paper is that image gradient is used to describe image patches. Specially, we define gradient image to efficiently use image gradient information for patch matching. By combining gradient image and limited PatchMatch, our basic gradient matching is more applicable to optical flow estimation than original PatchMatch. Further, we uncover the secrets of limited PatchMatch through a thorough analysis, and design an effective pyramidal matching framework based on these secrets. Our pyramidal matching framework is tailored to gradient image and able to find more inliers and reject outliers. Moreover, we prove that gradient image is not only effective and efficient but also scalable, which can provide a compromise between speed and accuracy. The evaluation on three modern datasets shows that our method is able to provide highly accurate optical flow on MPI Sintel and KITTI with competitive speed. Especially, our method outperforms state of the art published methods on both clean pass and final pass of MPI Sintel dataset. %In addition, the high efficiency of gradient image is also attractive.
In future, we are interested in improving our performance in homogeneous regions.
%============================================================================
\appendices
%\section*{Acknowledgment}
\section*{Acknowledgment}
This work is supported by the National Natural Science Foundation of China (Grant No. 61272043).

\ifCLASSOPTIONcaptionsoff
  \newpage
\fi



%\bibliographystyle{IEEEtran}
\bibliographystyle{IEEEtran}


%\begin{IEEEbiography}[{\includegraphics[width=1in,height=1.25in,clip,keepaspectratio]{author_1_new.pdf}}]{Yuanwei Li}
%received B.Sc. degree in Automation Science from Beijing University of Aeronautics \& Astronautics, and M.S. degree in Computer Science from National University of Defense Technology. His main research interests focus on pattern recognition, image processing and computer vision.
%\end{IEEEbiography}


%\vfill


\end{document}